\documentclass[lettersize,journal]{IEEEtran}
\usepackage{newtxtext,newtxmath}
\usepackage{booktabs,tabularx}
\usepackage{multirow}
\usepackage{subfigure}
\usepackage{amsmath,amsfonts}
\usepackage{algorithm}
\usepackage{tabularx}
\usepackage[noend]{algpseudocode}
\usepackage{array}
\usepackage[caption=false,font=normalsize,labelfont=sf,textfont=sf]{subfig}
\usepackage{textcomp}
\usepackage{stfloats}
\usepackage{url}
\usepackage{verbatim}
\usepackage{graphicx}
\usepackage{cite}
\usepackage{wasysym}
\usepackage[cmyk]{xcolor}
\begin{document}

\title{Large Language Models for Link Stealing Attacks Against Graph Neural Networks}

\author{Faqian Guan,~%\ead{faqianguan@gmail.com}
        Tianqing Zhu*,
        Hui Sun,~ %\ead{sunhuimk@gmail.com}
        Wanlei Zhou,~\IEEEmembership{Senior Member,~IEEE},~%\ead{wlzhou@cityu.edu.mo}
        and Philip S. Yu,~\IEEEmembership{Fellow,~IEEE,} %\ead{psyu@uic.edu}

\IEEEcompsocitemizethanks{\IEEEcompsocthanksitem Faqian Guan is with the China University of Geosciences, Wuhan, China; Tianqing Zhu, Hui Sun, and Wanlei Zhou are with the City University of Macau, Macau, China. Philip S. Yu is with University of Illinois Chicago, Chicago, IL, USA\protect\\

\IEEEcompsocthanksitem Tianqing Zhu is the corresponding author. E-mail: tqzhu@cityu.edu.mo}

\thanks{Manuscript received June 19, 2024; revised XX XX, 2024.}}

% The paper headers
\markboth{Journal of \LaTeX\ Class Files,~Vol.~14, No.~8, August~2021}%
{Shell \MakeLowercase{\textit{et al.}}: A Sample Article Using IEEEtran.cls for IEEE Journals}

\maketitle

\begin{abstract}

Graph data contains rich node features and unique edge information, which have been applied across various domains, such as citation networks or recommendation systems. Graph Neural Networks (GNNs) are specialized for handling such data and have shown impressive performance in many applications. However, GNNs may contain of sensitive information and susceptible to privacy attacks. For example, link stealing is a type of attack in which attackers infer whether two nodes are linked or not. Previous link stealing attacks primarily relied on posterior probabilities from the target GNN model, neglecting the significance of node features. Additionally, variations in node classes across different datasets lead to different dimensions of posterior probabilities. The handling of these varying data dimensions posed a challenge in using a single model to effectively conduct link stealing attacks on different datasets. To address these challenges, we introduce Large Language Models (LLMs) to perform link stealing attacks on GNNs. LLMs can effectively integrate textual features and exhibit strong generalizability, enabling attacks to handle diverse data dimensions across various datasets. We design two distinct LLM prompts to effectively combine textual features and posterior probabilities of graph nodes. Through these designed prompts, we fine-tune the LLM to adapt to the link stealing attack task. Furthermore, we fine-tune the LLM using multiple datasets and enable the LLM to learn features from different datasets simultaneously. Experimental results show that our approach significantly enhances the performance of existing link stealing attack tasks in both white-box and black-box scenarios. Our method can execute link stealing attacks across different datasets using only a single model, making link stealing attacks more applicable to real-world scenarios.

\end{abstract}


\section{Introduction}

\IEEEPARstart{G}{raphs} are powerful tools for modeling complex relationships between entities and are becoming increasingly common in today's data-driven environment. For instance, in citation networks \cite{DBLP:conf/iclr/KipfW17, zheng2021grb}, papers act as nodes and edges represent citation relationships. To address such graph data, researchers have proposed Graph Neural Networks (GNNs)  \cite{DBLP:journals/tnn/ScarselliGTHM09}. GNNs can handle the relationships between nodes and facilitate message passing, enabling effective processing of node-level \cite{DBLP:journals/tbd/XueZJT24}, link-level \cite{DBLP:journals/tbd/LiuZZF23}, and graph-level \cite{DBLP:conf/kdd/MaWYS23} tasks.

GNNs contain a wealth of sensitive information, such as graph data and model parameters. As GNN applications become increasingly prevalent, concerns about the privacy of GNNs are also growing \cite{DBLP:journals/air/GuanZZC24}. Privacy issues are prevalent in machine learning, and attackers exploit vulnerabilities to steal models, data, and other sensitive information \cite{DBLP:conf/uss/TramerZJRR16, DBLP:conf/sp/ShokriSSS17, DBLP:conf/uss/FredriksonLJLPR14}. One such attack, known as link stealing \cite{DBLP:conf/uss/HeJ0G021}, specifically targets GNNs. In this attack, attackers access service provider resources to illicitly obtain graph data links, resulting in privacy leaks.

\begin{figure}[t]%[htbp]
%是可选项 h表示的是here在这里插入，t表示的是在页面的顶部插入
\centering
\includegraphics[scale=0.46]{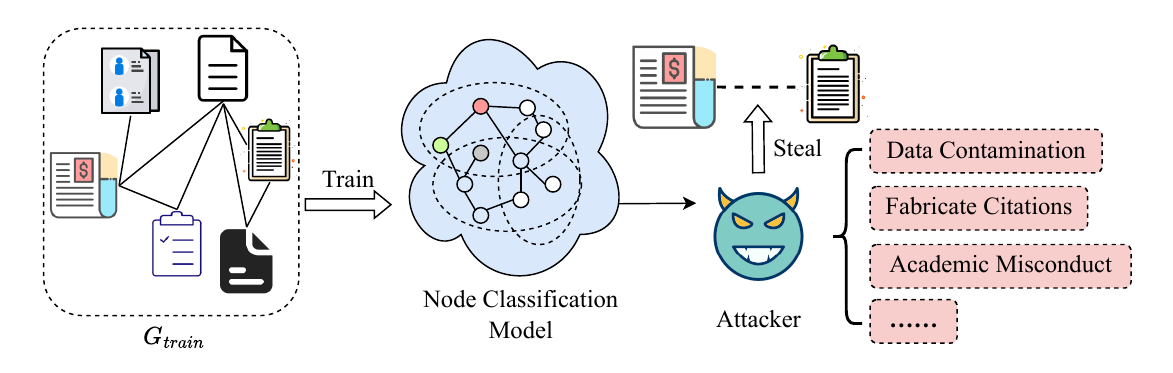}
\caption{Real-world instance of link stealing attacks in Graph Neural Networks.}
\label{realism}
\end{figure}

Fig. \ref{realism} illustrates a real-world example of a link stealing attack. The service provider employs the available citation dataset to train a GNN model for node classification, which is subsequently deployed online. Users can then access this GNN model to query the category to which a paper belongs. An attacker may leverage the response content obtained to infer the existence of a link between two papers. This link stealing attack poses a threat to the integrity of the citation dataset. By conducting this attack, attackers can manipulate citation relationships, compromising data accuracy and reliability. Furthermore, attackers might exploit the stolen link information to identify and replicate the citation patterns of high-impact papers or fabricate citations to artificially inflate the citation rate of a specific paper. Such actions can lead to academic misconduct, affecting academic reputations and the allocation of research grants.

Given the practical implications of link stealing attacks, many researchers have focused on investigating these attacks in GNNs \cite{DBLP:conf/uss/HeJ0G021,DBLP:conf/icml/ZhangWWYXPY23,DBLP:journals/corr/abs-2307-13548,DBLP:conf/sp/0011L0022}. In link stealing attacks, attackers obtain the node posterior probabilities through access to the target model. The mainstream approach of link stealing attacks is to compare the similarity between the posterior probabilities of different nodes to determine the presence of links \cite{DBLP:conf/uss/HeJ0G021, DBLP:conf/icml/ZhangWWYXPY23}. Furthermore, some researchers have inferred the presence of links by comparing changes in node posterior probabilities after perturbing the original graph data \cite{DBLP:journals/corr/abs-2307-13548,DBLP:conf/sp/0011L0022}. 

However, existing methods rely on posterior probabilities, ignoring the node features, especially textual features. Textual features have recently been used to improve the performance of GNN-related tasks \cite{DBLP:journals/corr/abs-2305-19523, DBLP:journals/natmi/LiuNWLQLTXA23}. We claim that node textual features can also enhance the accuracy of GNN privacy attacks. Furthermore, due to the different number of categories, different datasets have varying posterior probability dimensions. Here, \emph{dimension} refers to the length of the posterior probability vectors. The varying posterior probability dimensions make it difficult to train a model that can perform link stealing across different datasets. This limitation restricts real-world applications. Therefore, it is challenging to explore \emph{how textual features can be utilized to enhance link stealing attacks} and \emph{how to use a single model to perform link stealing attacks across diverse datasets}.

Recently, Large Language Models (LLMs) \cite{DBLP:journals/corr/abs-2303-18223} have achieved significant success in natural language processing (NLP) tasks \cite{DBLP:journals/corr/abs-2304-00228, DBLP:conf/iclr/ZengLDWL0YXZXTM23, DBLP:journals/corr/abs-2305-15268}. Given their superior performance, LLMs have been employed to address various machine learning challenges across different domains, such as data scarcity \cite{DBLP:conf/aaai/ChenXDWCZ024}, media bias detection \cite{DBLP:conf/eacl/LinWZLW24}, and recommender systems \cite{DBLP:conf/wsdm/LiuCS024}. Furthermore, some researchers have explored the use of LLMs to enhance downstream tasks in GNNs, achieving promising results \cite{DBLP:journals/corr/abs-2305-19523, DBLP:journals/natmi/LiuNWLQLTXA23, guo2024graphedit}. However, it remains to be explored how LLMs can utilize the textual features of the nodes to enhance the potential of privacy attacks, especially link stealing attacks in GNNs. Specifically, how to utilize these textual features to obtain a generalized LLM model that can simultaneously perform link stealing across different datasets.

In this paper, we introduce the use of LLMs to enhance link stealing attacks in GNNs. The LLM effectively combines textual features and posterior probabilities of nodes, significantly improving the performance of the attacks. Depending on the attacker's knowledge, we perform attacks in both white-box and black-box settings. To tailor the LLM for the link stealing attack task, we design two distinct prompts specifically for fine-tuning in these settings. These prompts effectively incorporate both textual features and posterior probabilities of nodes. By fine-tuning the LLM with these prompts, we enable the model to adapt effectively to the link stealing task.

Furthermore, we can use a single LLM to perform link stealing attacks across diverse datasets. LLMs have a strong generalization ability and can handle variable-length textual features \cite{DBLP:journals/corr/abs-2303-18223}. By leveraging this capability, we fine-tune the LLM using multiple datasets with our designed prompts. This process allows the LLM to learn features from various datasets, enhancing its inference ability across them. As a result, we obtain a versatile LLM capable of simultaneously conducting link stealing attacks on diverse datasets. Experimental results demonstrate that our method significantly improves the performance of link stealing attacks. Additionally, our approach enables link stealing across different datasets using a single LLM model, showcasing its practical importance.

In summary, our paper contributes in the following:

\begin{itemize}

    \item We introduce LLMs to enhance the efficiency of link stealing attacks on GNNs, resulting a more realistic attacks in a real-world scenario.

    \item We develop two distinct LLM prompts to effectively combine textual features and posterior probabilities of graph nodes in both black-box and white-box scenarios, enhancing link stealing attacks.

    \item With our method, we can carry out link stealing attacks across different graph datasets using just one LLM, achieving better results and making the attacks more realistic.

\end{itemize}

\section{Related work}
\subsection{Privacy Attacks}
Privacy attacks on machine learning models involve the unauthorized acquisition of information about a model and its training data through accessing the deployed model by service providers. Tramèr et al. \cite{DBLP:conf/uss/TramerZJRR16} introduce model extraction attacks, where an attacker can replicate a similar or identical machine learning model by analyzing the data and corresponding responses. Shokri et al. \cite{DBLP:conf/sp/ShokriSSS17} propose membership inference attacks to determine whether data are in a model's training dataset by sending data to access the model. Fredrikson et al. \cite{DBLP:conf/uss/FredriksonLJLPR14} study attribute inference attacks, which exploit available data and information from the service provider to extract private data from the model.

Graph Neural Networks (GNNs), a specialized machine learning model, have attracted attention regarding privacy attacks \cite{DBLP:journals/air/GuanZZC24}.\cite{DBLP:conf/asiaccs/WuYPY22,DBLP:conf/sp/ShenHH022,GUAN2024112144} examine model extraction attacks on GNNs, with the aim of pilfering the service provider's model under varying levels of attack knowledge. \cite{DBLP:journals/corr/abs-2102-05429,DBLP:conf/icdm/WuYPY21} explore membership inference attacks on GNNs to ascertain node presence in the target model's training set. Zhang et al. \cite{zhang2022inference} introduce a unique form of membership inference attack, determining if subgraphs belong to the entire graph. Furthermore, Li et al. \cite{DBLP:conf/cikm/LiLL020} delve into privacy inference attacks on GNNs to deduce attributes of the target graph. Their research on de-anonymization includes presenting a seed-free graph de-anonymization method that automatically extracts features and performs global node matching without initially matching node pairs. Zhang et al. \cite{zhang2022inference} also examine attribute inference attacks via graph embeddings, aiming to infer essential attributes of the target graph, such as the count of nodes, the number of edges, and the graph density.

\textbf{Link Stealing Attacks.} 
Link stealing attacks pose a significant privacy threat to GNNs, targeting the specific attribute features of graph data known as links. Attackers leverage the service provider's GNN model to obtain the posterior probability of a node. By combining this probability with node information, they infer link existence between nodes. He et al. \cite{DBLP:conf/uss/HeJ0G021} introduced link stealing attacks and explored them under eight different levels of attacker knowledge. They proposed various novel approaches for these eight attacks. Extensive experiments demonstrated the effectiveness of their methods in stealing links. Zhang et al. \cite{DBLP:conf/icml/ZhangWWYXPY23} provided theoretical evidence for the uneven vulnerability of GNNs to link stealing attacks, laying the groundwork for understanding the varying risks among different edge groups. They also introduced a group-based attack paradigm to highlight the actual link stealing hazards faced by GNN users due to this uneven vulnerability. Wu et al. \cite{DBLP:conf/sp/0011L0022} proposed a privacy attack via impact analysis, using adversarial queries to infer private edge information. Zari et al. \cite{DBLP:journals/corr/abs-2307-13548} inferred private edge information in graph-structured data by injecting new nodes and influencing the posterior probabilities of node predictions to expose privacy vulnerabilities in GNNs. 

While previous methods have proven effective, they are relatively simple, often relying on similarity or multilayer perceptron (MLP) \cite{rosenblatt1958perceptron} models and not fully incorporating node features. Moreover, due to differences in data dimensions across datasets, previous methods can struggle to conduct link stealing attacks on various datasets. To address these limitations, this paper leverages LLMs, combined with node textual features, to enhance the performance of link stealing attacks and enable the use of a single LLM to perform link stealing attacks across different datasets.

\subsection{Large Language Models}
Large Language Models (LLMs) \cite{DBLP:journals/corr/abs-2303-18223} are gaining traction in both academic and industrial domains. Various LLMs have been proposed, including GPT \cite{DBLP:journals/mima/FloridiC20}, Vicuna \cite{DBLP:conf/nips/ZhengC00WZL0LXZ23} and LongChat \cite{longchat2023}. Initially designed for natural language processing tasks, researchers have explored their applications in different areas. Yang et al. \cite{DBLP:journals/corr/abs-2304-00228} investigated text categorization using LLMs, while Zeng et al. \cite{DBLP:conf/iclr/ZengLDWL0YXZXTM23} explored LLMs for sentiment analysis. Tao et al. \cite{DBLP:journals/corr/abs-2305-15268} conducted a comprehensive evaluation of LLMs' event semantic processing capabilities, revealing strengths in understanding individual events but limitations in perceiving semantic similarities between events.

LLMs are highly effective and have expanded into various fields. Chen et al. \cite{DBLP:conf/aaai/ChenXDWCZ024} tackled the challenge of limited data resources by using artificial data generated by LLMs. This approach helps address the issue of data scarcity, especially in cross-modal applications, improving their performance and usability. Lin et al. \cite{DBLP:conf/eacl/LinWZLW24} explored LLMs for media bias detection, addressing limitations of previous methods that relied on specific models and datasets, which hindered adaptation and performance on out-of-domain data. Liu et al. \cite{DBLP:conf/wsdm/LiuCS024} explored the use of LLMs in recommender systems, enhancing content-based recommendation approaches with both open-source and closed-source LLMs.

\textbf{Large Language Models on Graph.} In recent times, LLMs have emerged as powerful tools for graph-related tasks, outperforming traditional GNN approaches and achieving state-of-the-art results. He et al. \cite{DBLP:journals/corr/abs-2305-19523} leverage semantic knowledge generated by LLMs to enhance the quality of initial node embeddings in GNNs. Ye et al. \cite{DBLP:journals/corr/abs-2308-07134} substitute the predictor in GNNs with LLMs to enhance node embeddings by flattening graphs and employing instruction prompts to leverage the expressive nature of natural language. Liu et al. \cite{DBLP:journals/natmi/LiuNWLQLTXA23} align GNNs and LLMs into a shared vector space and integrate textual knowledge into graphs to enhance inference capabilities. Guo et al. \cite{guo2024graphedit} propose GraphEdit, which leverages LLMs to learn complex node relationships in graph-structured data. This approach overcomes the limitations associated with explicit graph structural information and enhances the reliability of graph structure learning.

Previous studies have shown that LLMs can effectively complement GNNs, enhancing various GNN-related tasks. However, exploring how LLMs can be utilized to intensify privacy attacks on GNNs, particularly in link stealing attacks, represents a novel direction for research. This paper aims to investigate how LLMs can be integrated to enhance the effectiveness of link stealing attacks on GNNs.

\section{Preliminary}
\subsection{Notations}
A graph dataset is denoted as $\mathrm{G}$. $\mathrm{X}$ represents the node features in $\mathrm{G}$, including numerical features $\mathrm{N}$ and textual features $\mathrm{T}$. $\mathrm{P}$ denotes the posterior probabilities of the nodes. $\mathrm{H}$ represents the hidden state of the feature vector, which serves as the node feature at the model's network layer. $u$ and $v$ represent individual nodes of the graph, with $\mathcal{N}(v)$ denoting the neighboring nodes of node $v$. $\mathcal{T}$ denotes the target GNN model that the attacker accesses to obtain $\mathrm{P}$. $\mathcal{A}$ denotes the attack model. The attacker uses $\mathcal{A}$ for link stealing attacks. $\mathrm{Y}$ and $\hat{\mathrm{Y}}$ denote the true labels and predicted labels of the nodes. The notations used in this paper are summarized in Table \ref{Notations}. 

\begin{table}[htbp]
  \centering
  \caption{Summary of notations.}
  
  \scalebox{1}{
    \begin{tabular}{c|c}
    \toprule
    \textbf{Notations} & \textbf{ Description} \\
    \midrule
    $\mathrm{G}$ & Graph data \\
    $\mathrm{X(N, T)}$ & Node features \\
    $\mathrm{N}$ & Numerical features of nodes \\
    $\mathrm{T}$ & Textual features of nodes \\
    $\mathrm{P}$ & Posterior probabilities of nodes \\
    $\mathrm{H}$ & Hidden state of a feature vector \\
    $u, v$ & Individual nodes in the graph \\
    $\mathcal{N}(v)$ & Neighboring nodes of node $v$ \\
    $\mathcal{T}$ & Target model \\
    $\mathcal{A}$ & Attack model \\
    $\mathrm{Y}$ & True labels of nodes \\
    $\hat{\mathrm{Y}}$ & Predicted labels of nodes \\
    $\mathrm{n, m}$ & Number of node pairs in the graph \\
    \bottomrule
    \end{tabular}%
    }
  \label{Notations}%
\end{table}%

\subsection{Graph Neural Networks}
Graph Neural Networks (GNNs) \cite{DBLP:journals/tnn/ScarselliGTHM09} are a type of neural network specifically designed to process input structured as a graph. They excel in tasks involving nodes and edges with complex interactions. GNNs aim to learn representations of nodes in a graph by aggregating information from their neighboring nodes. This process is commonly achieved through message passing, where each node receives and integrates data from its neighboring nodes, updates its own representation based on this integrated data, and subsequently communicates this updated representation to its neighboring nodes in the next iteration. The updated representation of a node at the $k$-th layer of the GNN can be expressed as follows:

\begin{equation} \label{gnn_eq}
    \begin{gathered}
        {h}_{v}^{(k)}=\operatorname{UPDATE}^{(k)}\left({h}_{v}^{(k-1)}, {e}_{v}^{(k)}\right) \\
        {e}_{v}^{(k)} = \operatorname {AGG}^{(k-1)}\left(\left\{{h}_{u}^{(k-1)}: \forall u \in \mathcal{N}(v) \cup v\right\}\right)    
        % h_i^{(k)}=\operatorname{UPDATE}\left(h_i^{(k-1)}, e_{j}^{(k-1)}\right)\\
        % e_{j}^{(k-1)} = \operatorname{AGGREGATE}\left(\left\{h_j^{(k-1)} \mid v_j \in \mathcal{N}\left(v_i\right)\right\}\right)
    \end{gathered}
\end{equation}
where ${h}_{v}^{(k)}$ denotes the updated representation of node $v$ at the $k$-th layer, and ${h}_{v}^{(0)}$ represents the initial input feature of node $v$ (i.e., $x_v$). $\mathcal{N}\left(v\right)$ signifies the neighbors of node $v$. $\operatorname{AGG(\cdot)}$ is an aggregation function that combines information from neighboring nodes. Additionally, $\operatorname{UPDATE(\cdot)}$ functions as an update function that integrates the aggregated information into the node's representation.

\begin{figure}[tp]%[htbp]
%是可选项 h表示的是here在这里插入，t表示的是在页面的顶部插入
\centering
\includegraphics[scale=0.48]{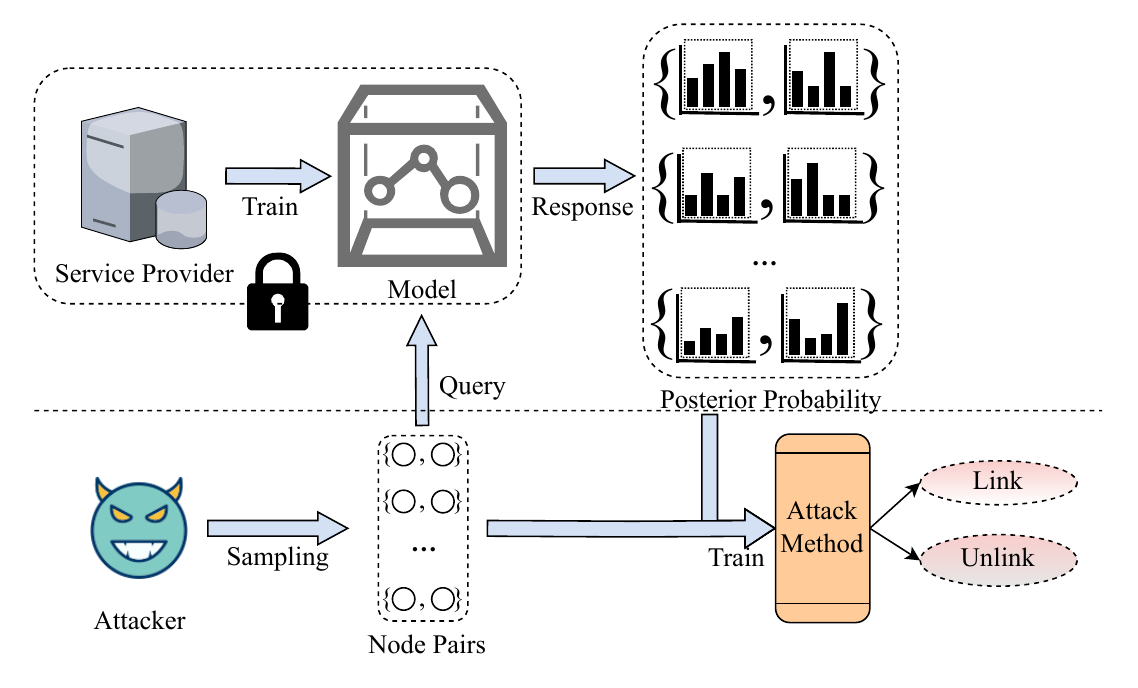}
\caption{Basic flow of link stealing attacks. The service provider trains the model and deploys it on the web for queries. The attacker queries the model using existing knowledge to obtain posterior probabilities. The attacker then conducts an attack using the obtained posterior probabilities and the original knowledge.}
\label{Basic}
\end{figure}

\subsection{Link Stealing Attacks}
Link Stealing Attacks refer to malicious activities where an attacker attempts to obtain the links within graph data used to train a GNN. This involves stealing the connections or relationships between nodes in the training graph data. These attacks exploit the model's predictions to extract sensitive or proprietary information about the underlying graph structure. Consequently, these attacks pose significant risks to the privacy and security of sensitive graph data.

The process of a link stealing attack is depicted in Fig. \ref{Basic}. Firstly, the service provider trains a GNN model for user use. This GNN model is called the target model in the attack, also known as the victim model. The attacker conducts the attack by accessing the target model. Secondly, the attacker samples different nodes that they intend to target to determine their connectivity, forming node pairs. Each node pair contains information about two different nodes. In the link stealing attack, the attacker directly assesses the node pair to infer whether there is a connection between the two nodes. Subsequently, the attacker queries the target model to obtain the posterior probabilities of these nodes. Armed with this information, the attacker explores suitable attack methods based on the node pairs and their respective probabilities. Finally, employing the devised attack method, the attacker determines whether the node pairs are linked, thus posing a significant threat to the privacy of the service provider.

\subsection{Large Language Models}
Large Language Models (LLMs) are advanced natural language processing models based on deep learning. They learn the meaning and structure of human language to generate text that reads like it's written by a human. A language model is a type of artificial intelligence that specializes in processing and understanding language. It identifies patterns in text data and generates new content based on those patterns when given prompts.

LLMs are typically built using neural network models and trained on huge datasets, like the vast amount of text found on the Internet. These models contain billions to trillions of parameters, which are like the building blocks that help them understand and process language. They can perform various natural language processing tasks effectively.

\textbf{Large Language Models in Graphs.} LLMs can assist GNNs with graph data, falling into two main categories \cite{DBLP:journals/corr/abs-2307-03393}:

\begin{itemize}

    \item \textbf{LLMs as enhancers:} These models leverage the vast knowledge of LLMs to improve the textual attributes of nodes. This enhancement aids GNNs in making predictions.

    \item \textbf{LLMs as predictors:} In this approach, LLMs are directly employed for predicting tasks related to graph data.
    
\end{itemize}

In this paper, we use LLMs as predictors to improve link stealing attacks in GNNs.

\textbf{Prompt in Large Language Models.} In the context of LLMs, a \emph{prompt} refers to a specific input or query provided to the model to generate a desired output. Prompts are structured in a way that guides the LLM to produce text or predictions that align with the intended task or query. They are crucial in directing the model's behavior and can be designed to elicit responses that are relevant to a particular application or problem domain.

In a link stealing attack, the prompt provided to the LLM may involve querying the model about information regarding two nodes, including textual data. The prompt specifies the format of the input data and the expected output format, ensuring that the LLM generates predictions that enable the inference of private information about the GNN's structure. Therefore, designing an appropriate prompt is crucial for the success of the link stealing attack.

\begin{figure*}[t]%[htbp]
%是可选项 h表示的是here在这里插入，t表示的是在页面的顶部插入
\centering
\includegraphics[scale=0.69]{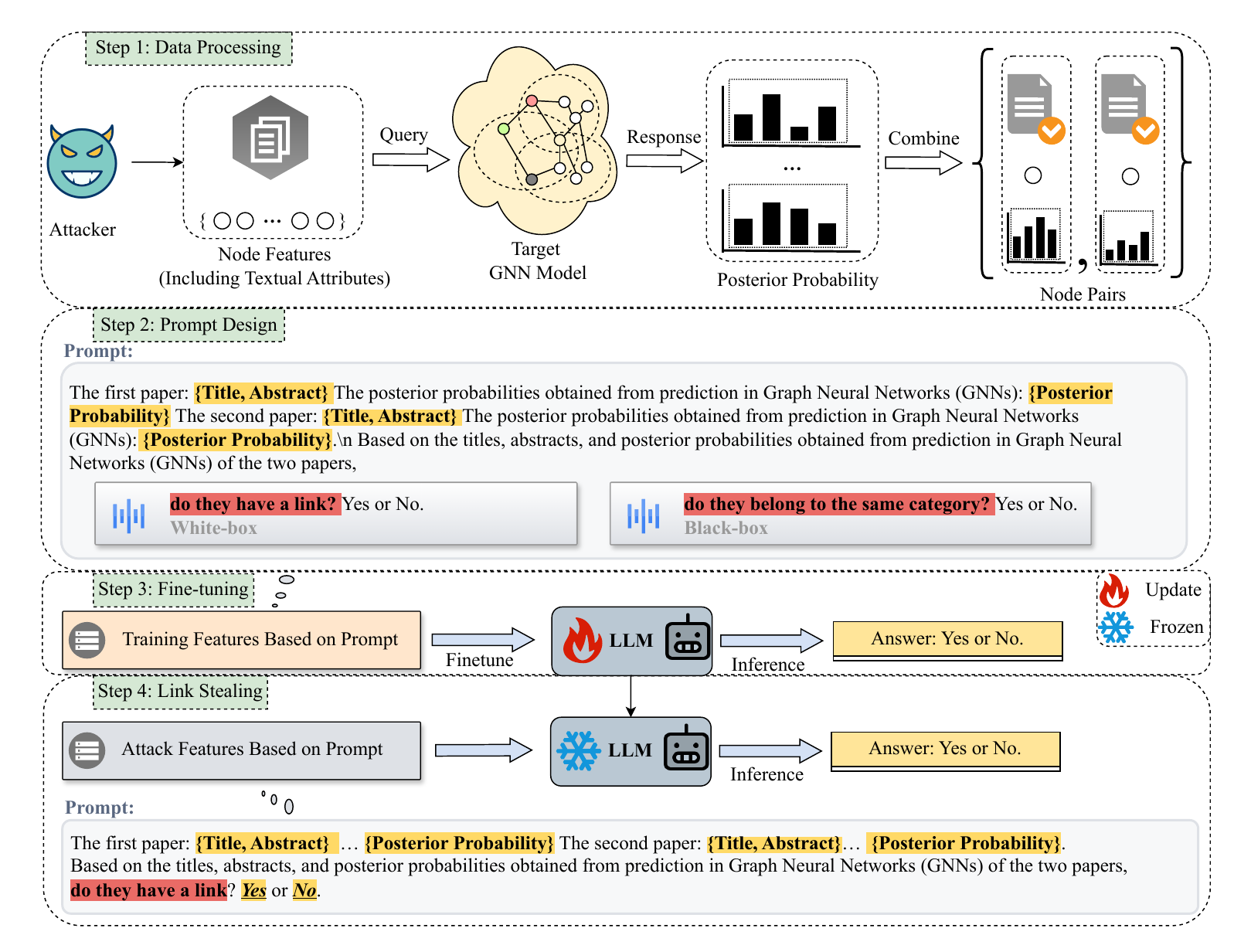}
\caption{Overview of proposed link stealing attack method. The Data Processing step involves creating node pairs containing features and posterior probabilities of the nodes. These pairs serve as input features for constructing the LLM attack model. In the Prompt Design step, different prompts are created for white-box and black-box settings. These prompts include information about the node pairs. In the Fine-tuning step, the LLM is fine-tuned using the designed prompts. In the Link Stealing step, the fine-tuned LLM is used to determine whether there is a connection between the node pairs.}
\label{Overview}
\end{figure*}

\section{LLMs-based Link Stealing Attacks}

\subsection{Threat Model}

\textbf{Problem Formulation.} The goal of the link stealing attack is to infer whether two nodes in a graph have a link or not. Specifically, a service provider trains a GNN model $\mathcal{T}$ and deploys it on the internet. Through access to the GNN model $\mathcal{T}$, the attacker can use attack model $\mathcal{A}$ to infer whether there exists a link between nodes $v$ and $u$. This attack can be formally defined as follows:

\begin{equation}
    \begin{gathered}
        \mathcal{A}: \{v, u\}, \mathcal{T} \mapsto \text ( \emph{Link}, \emph{Unlink} )
    \end{gathered}
\end{equation}
The attacker inputs nodes $v$ and $u$ into $\mathcal{T}$ to obtain their posterior probabilities. The attacker then combines the node features and posterior probabilities, using $\mathcal{A}$ to determine whether there is a link between nodes $u$ and $v$. \emph{Link} indicates the presence of a connection, while \emph{Unlink} indicates its absence.

\textbf{Target Model.} In a link stealing attack, the target model is $\mathcal{T}$. The attacker's goal is to obtain the link information from the training data of the target model through various attack methods, thereby compromising privacy security and potentially facilitating illegal exploitation.

\textbf{Adversarial Knowledge.} In privacy attacks, attackers steal private information with varying levels of knowledge. In this paper, we investigate link stealing attacks in both white-box and black-box scenarios.

In white-box scenarios, the attacker has knowledge about the links of some nodes and node features. Except for these, the attacker has no knowledge of other information, such as the parameters and training process of the target model. This aligns with real-world situations where service providers must safeguard the privacy of their models, thereby enhancing the realism of link stealing attacks.

In black-box scenarios, the attacker has neither knowledge about the links between nodes, nor information about the target model's parameters and training process. In this scenario, the attacker conducts link stealing attacks solely based on node features. This type of attack demands minimal prior knowledge. In reality, attackers can easily acquire the necessary knowledge to conduct black-box attacks.

Theoretically, the white-box attacker can infer links between nodes more accurately. Next, we explore how LLMs can enhance attacks in these two scenarios.

\subsection{Attack Overview}
In the link attack, LLMs effectively combine textual features and posterior probabilities of nodes to enhance the performance. Meanwhile, by incorporating LLMs, we leverage their strong generalization capabilities, enabling a single model to effectively conduct link stealing tasks across varied datasets, thereby enhancing the realism of link stealing attacks. Our method, outlined in Fig. \ref{Overview}, consists of four main steps: Data Processing, Prompt Design, Fine-tuning, and Link Stealing. 

In the Data Processing step, the attacker creates node pairs that contain the features and posterior probabilities of the nodes. These posterior probabilities are obtained by querying the target GNN model. The attacker uses these node pairs as features for constructing the attack model, which can then determine if two nodes in a pair are linked. When using LLMs for link stealing attacks, the attacker needs to design prompts that include information about the node pairs to fine-tune the LLM. In the Prompt Design step, given that the attacker has more information about the links between node pairs in a white-box setting compared to a black-box setting, we design two different prompts corresponding to these scenarios. In the Fine-tuning step, the attacker fine-tunes the LLM using the designed prompts. In the Link Stealing step, the attacker uses the fine-tuned LLM to determine whether there is a connection between the node pairs. It is worth mentioning that in the Link Stealing step, regardless of whether it is a white-box or black-box setting, the attacker uses the same prompt for link inference, as illustrated in Fig. \ref{Overview}.

\subsection{Data Processing}
During the data processing phase of the attack, the attacker first accesses the target model using node features to obtain the posterior probabilities of node classification. In link stealing attacks, the attacker aims to determine whether two nodes have a link. To facilitate this determination, we combine two nodes into a node pair during data processing. This node pair includes the features of both nodes along with their posterior probabilities. We then determine whether they have a link by assessing the node pair directly, as follows:

\begin{equation}
    \begin{gathered}
        \mathcal{A}: \underbrace{\{(x_v, p_v), (x_u, p_u)\}}_\mathrm{Node\ Pair},\mapsto \text ( \emph{Link}, \emph{Unlink} )
    \end{gathered}
\end{equation}
where $x_v$ and $x_u$ represent the node features of nodes $v$ and $u$, including textual descriptions, and $p_v$ and $p_u$ represent their posterior probabilities. The attacker combines $x_v$ and $p_v$ of node $v$ with $x_u$ and $p_u$ of node $u$ to form a node pair. The attacker directly inputs the node pair into the attack model to determine whether there is a link between $u$ and $v$ in the node pair. \emph{Link} indicates that there is a link existed, and \emph{Unlink} indicates no link existing.

\subsection{Prompt Design}
\begin{figure}[htp]%[htbp]
%是可选项 h表示的是here在这里插入，t表示的是在页面的顶部插入
\centering
\includegraphics[scale=0.8]{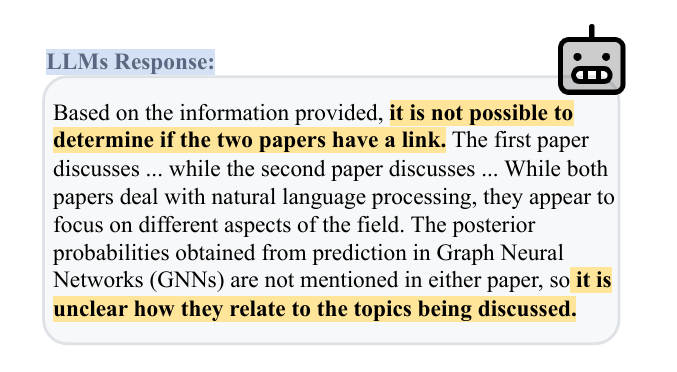}
\caption{Response to link stealing attacks by the original LLM model in the black-box setting.}
\label{black-example}
\end{figure}

LLMs have good performance in many tasks. However, LLMs' performance in some specialized downstream tasks are not optimal. For instance, in link stealing attacks, the original LLM's performance is notably poor, to the extent that LLMs cannot effectively complete the link stealing attack task. The original LLM's response to link stealing attacks is illustrated in Fig. \ref{black-example}, where the LLM fails to accurately predict whether there is a link between two nodes.

To enhance the performance of the link stealing attack, the original LLM requires adaptation. Task-specific prompt fine-tuning is a method to adapt LLMs for specific tasks by integrating task-specific prompts \cite{DBLP:journals/csur/LiuYFJHN23}.

In this step, we employ prompt fine-tuning to enhance LLM's performance on the downstream task: link stealing attack. The prompts consist of three components: i) node pair information, ii) human question, and iii) the LLM's response. Node pair information includes descriptions of node features in natural language and the probabilities of node classifications obtained from target GNN models. The human question serves as the natural language input for the LLM. The LLM's response is the answer to the human question. 

\begin{figure}[htp]%[htbp]
\centering
\includegraphics[scale=0.55]{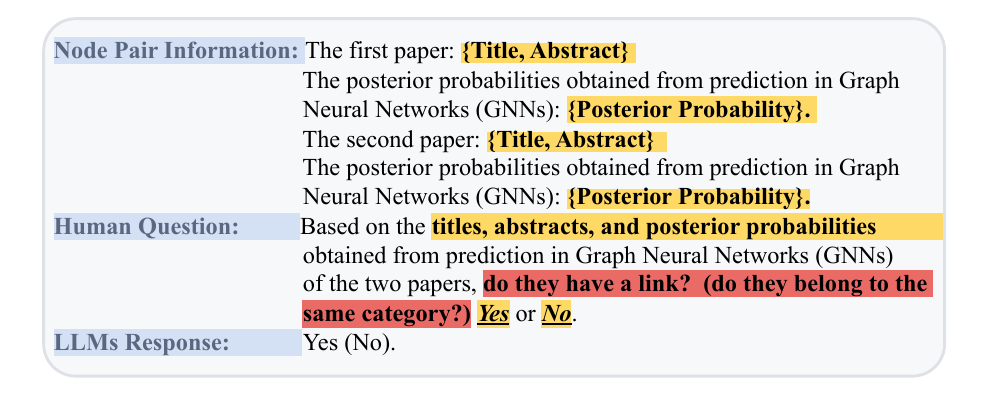}
\caption{Our prompt designs for link stealing attacks.}
\label{prompt}
\end{figure}

An example of the prompt data for the link stealing attack is illustrated in Fig. \ref{prompt}. This prompt is designed based on a citation dataset. The node pair information contains the titles and abstracts of two papers, as well as the posterior probabilities of the nodes obtained by accessing the target GNN. In the human question, based on the information in the node pair, the LLM is asked whether there is a link between these two papers (whether they belong to the same category). In the LLM's response, the LLM answers the human question. \emph{Yes} indicates that the two papers have a link (same category), and \emph{No} indicates that there is no link (different category).

Due to differences in the attacker's knowledge, it's crucial to highlight the variation in the prompt design of the human question in our approach.

\begin{itemize}

    \item \textbf{White-box Setting:} In white-box attacks, the attacker knows whether links exist for certain nodes. Consequently, we can extract node pairs involving these nodes through data processing. We then fine-tune the LLM for the link stealing task using these pairs. Specifically, we set the Human Question: \emph{do they have a link?} This allows us to adapt the LLM to the link stealing task.
    
    \item \textbf{Black-box Setting:} In black-box attacks, the attacker lacks knowledge of links between nodes. To assist the LLM in performing link stealing tasks on GNNs, we fine-tune it using shadow tasks. These shadow tasks are other GNN-related tasks that are related to the link stealing task. By training on these shadow tasks, the LLM can generate responses that are more effective for the link stealing task.

    In graph data, nodes of the same class are often more likely to have a link than nodes of different classes \cite{DBLP:journals/corr/abs-2402-15183}. This implies a strong correlation between node same-class inference and link inference tasks. Therefore, the node same-class inference task serves as a shadow task for the link stealing attack. In the node same-class inference task, the model determines whether two nodes belong to the same class. Hence, we fine-tune the LLM on this task by asking the question: \emph{Do they belong to the same category?} This related task assists the model in determining if two nodes are from the same class, thereby facilitating the LLM in predicting the presence of a link between them.

\end{itemize}

\subsection{Fine-tuning}
\begin{figure}[htp]%[htbp]
%是可选项 h表示的是here在这里插入，t表示的是在页面的顶部插入
\centering
\includegraphics[scale=0.53]{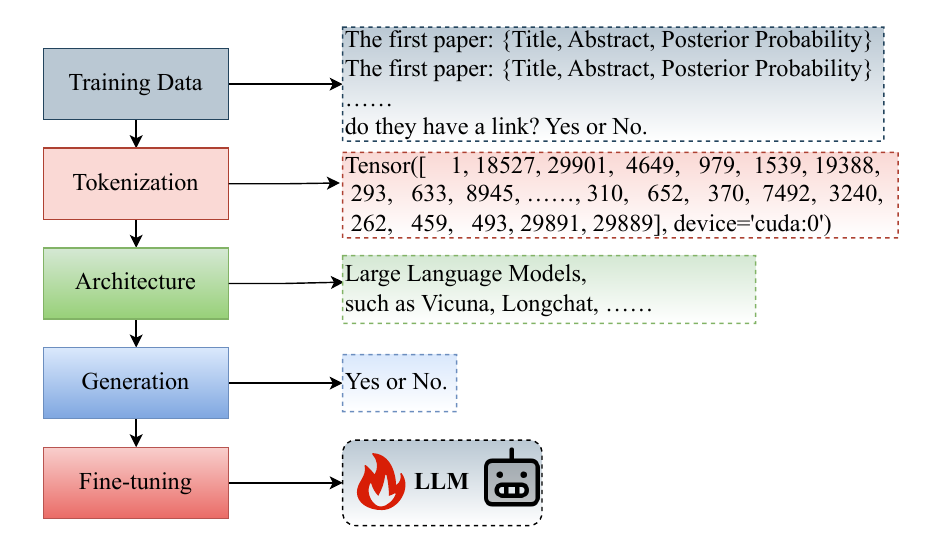}
\caption{Steps to train LLM for link stealing attacks.}
\label{Finetune}
\end{figure}

This process involves adjusting the LLM with task-specific graph prompts, aiding the model in generating responses more suitable for the graph learning task. This enhances the model's adaptability and improves its performance in addressing graph link stealing attacks. Specifically, as depicted in Fig. \ref{Finetune}, we break down the fine-tuning LLM process into five parts for a detailed presentation:

\begin{itemize}

    \item \textbf{Training Data:} This refers to the dataset used to fine-tune the LLM. It typically consists of a large amount of text data. In our attack method, it refers to the prompts we designed, which include textual features and the posterior probabilities of the node pairs.
    
    \item \textbf{Tokenization:} Tokenization is the process of breaking down text into smaller units known as tokens \cite{DBLP:conf/acl/SennrichHB16a}. Tokens can be words, subwords, or characters. Each token is assigned a unique numerical representation. Tokenization aids in the model's ability to process and comprehend text more effectively.

    \item \textbf{Architecture:} The language model usually comprises several layers of transformer neural networks. These networks are robust models capable of analyzing and capturing relationships between various tokens in the text.

    \item \textbf{Generation:} During generation, the model receives a prompt or question and uses its learned knowledge to produce a sequence of tokens that fits the given context. In our approach, the LLM generates \emph{Yes} or \emph{No} responses depending on the prompt.

    \item \textbf{Fine-tuning:} During fine-tuning, the model's parameters are adjusted based on the task's objectives and the training data. In our method, the LLM generates content $\hat{y}$ from the prompt, which the attacker then compares with real content $y$ to calculate the loss. Some of the model's parameters are fine-tuned based on this loss. We calculate the loss using cross-entropy $\mathcal{L}_{C E}$ as the loss function, represented as:

    \begin{equation}
        \begin{gathered}
            \hat{y} = \operatorname{LLM}(\operatorname{Prompt}\{(x_v, p_v), (x_u, p_u)\})\\
            \mathcal{L}_{C E}\left(y, \hat{y}\right) =-\left[y \operatorname { l o g } (\hat{y}+\left(1-y\right) \log \left(1-\hat{y}\right)\right]
        \end{gathered}
    \end{equation}

\end{itemize}

It is worth noting that LLMs, due to their superior generalization ability, can learn from graph data across different domains while maintaining strong performance in all of these domains \cite{DBLP:journals/corr/abs-2310-13023}. Previous link stealing attack methods focus solely on posterior probabilities. Because of the varying dimensions of posterior probabilities across different domains, it is difficult to train a generalized model that fits multiple domain data. However, LLMs handle variable-length text data effectively, even when dealing with graph data of different dimensions.

When fine-tuning the LLM, we input prompts from datasets of various domains to train it. This enables the LLM to learn information from diverse datasets, improving its generality. Consequently, we obtain a generalized LLM capable of performing link stealing attacks across different datasets, making these attacks more applicable to real-world scenarios.

\subsection{Link Stealing}
By using a fine-tuned LLM model, the attacker can execute link stealing attacks. Initially, the attacker creates node pairs comprising two nodes targeted for link stealing. Subsequently, the attacker accesses the target GNN model to retrieve the posterior probabilities of the two nodes. Then, the node features and posterior probabilities are combined using well-designed prompts to create attack features. Finally, the attack features are inputted into the fine-tuned LLM model to inquire about the presence of links between the node pairs.

We employ different prompt designs for fine-tuning LLMs in black-box and white-box settings, reflecting the variability in the attacker's adversarial knowledge. However, in the link stealing stage, as depicted in Fig. \ref{Overview}, we utilize the same prompt in both black-box and white-box designs, namely, \emph{do they have a link?} This prompt enables the attacker to execute the link stealing attack effectively.

Algorithm \ref{Link Stealing Attacks} outlines the link stealing attack process. Line $1$ involves data processing, where node pairs are formed using the graph dataset $\mathrm{G}$, which includes nodes from different domains. In line $2$, the attacker accesses the target GNN model to acquire the posterior probability $\mathrm{P}$ of each node. Line $3$ entails prompt design, while lines $4$ to $6$ denote the fine-tuning of the LLM using the designed prompts. Line $7$ signifies the link stealing attack conducted by the fine-tuned LLM model.

\begin{algorithm}[ht]
\caption{The Process of Link Stealing Attacks.}\label{Link Stealing Attacks} %算法的名字
\hspace*{0.02in} {\bf Input:} %算法的输入， \hspace*{0.02in}用来控制位置，同时利用 \\ 进行换行
$\mathrm{G}$, and $\mathcal{T}$, $\operatorname{LLM}$\\
\hspace*{0.02in} {\bf Output:} %算法的结果输出
$Link\ or\ Unlink$
\begin{algorithmic}[1]

\State Sampling $\mathrm{G}$ $\rightarrow$ creating node pairs, such as $\{v, u\}$.
\State Getting node posterior probabilities $\rightarrow$ $\mathrm{P} = \mathcal{T}(\mathrm{X})$.
\State Designing Prompt $\rightarrow$  $\operatorname{Prompt}\{(x_v, p_v), (x_u, p_u)\}$.
\Repeat 
    \State Fine-tuning the LLM model using prompts.
\Until convergence  
% \State Link stealing by fine-tuned LLM model and Prompt
\State $Link\ or\ Unlink = \operatorname{LLM}(\operatorname{Prompt}\{(x_v, p_v), (x_u, p_u)\})$ \\
\Return $Link\ or\ Unlink$

\end{algorithmic}
\end{algorithm}

\subsection{Theoretical Analysis}
Here, we propose theoretical proof about using LLMs for link stealing attacks against GNNs.

\subsubsection{Analysis that LLMs enhance the performance of attacks} 
LLMs can effectively utilize the textual features of nodes to enhance the performance of link stealing attacks. In previous link stealing attacks, due to methodological limitations, researchers did not utilize the textual structural features of the nodes. Researchers only constructed the attack features $\mathrm{X_1}$ using the numerical features $\mathrm{M}$ and the posterior probabilities $\mathrm{P}$ of the nodes. When constructing the LLM attack model, we use the numerical features $\mathrm{M}$, the posterior probabilities $\mathrm{P}$, and the unique textual features $\mathrm{T}$ of the nodes as the training features $\mathrm{X_2}$ of the attack model. Clearly, we can infer that:

\begin{equation}
    \begin{gathered}
        \mathrm{X_1(M,P)} \subset \mathrm{X_2(M,P,T)}
    \end{gathered}
\end{equation}

Meanwhile, let $\mathcal{A}$ represent an attack model. Let $\mathcal{A}(\mathrm{X_1})$ represent the probability of a successful link stealing attack using $\mathrm{X_1}$, and $\mathcal{A}(\mathrm{X_2})$ represent the probability of a successful link stealing attack using $\mathrm{X_2}$. Since $\mathrm{X_1}$ includes more features, it is reasonable to assume that there is a positive gain coefficient $\alpha (\alpha > 1)$, making $\mathcal{A}(\mathrm{X_2}) = \alpha \mathcal{A}(\mathrm{X_1})$. This coefficient represents the enhancement in performance due to the additional features provided by the LLM.

Furthermore, with the right prompt, LLM can be ensured that $\operatorname{LLM}(\operatorname{Prompt}(\mathrm{X_1}) > \operatorname{Pre}(\mathrm{X_1})$ in the context of a link stealing attack, where $\operatorname{Pre}$ represents the previously common method. To sum up, with an appropriate prompt, we can infer that:

\begin{equation}
    \begin{gathered}
        \operatorname{LLM}(\operatorname{Prompt}(\mathrm{X_2}) >> \operatorname{Pre}(\mathrm{X_1})
    \end{gathered}
\end{equation}

Based on the above analysis and assumptions, it is shown that LLMs can be utilized for link stealing attacks against GNNs under appropriate prompts. 

\subsubsection{Analysis that a single LLM performs attacks across different datasets} This paper introduces LLMs for link stealing attacks, addressing the limitation in previous research where a single model could not perform attacks across different datasets. Additionally, using multiple datasets to train a single attack model can enhance the number of training samples and improve the performance of the attack model.

Suppose there are two datasets, $\mathrm{G_1}$ and $\mathrm{G_2}$. $\mathrm{G_1}$ has $3$ node classes, and $\mathrm{G_2}$ has $6$ node classes. Because of the different node classes, the dimensions of the posterior probabilities are also different. For instance, $p_1 = [0.15, 0.72, 0.13]$, and $p_2 = [0.05, 0.07, 0.08, 0.12, 0.58, 0.10]$. These are the posterior probabilities of the nodes in $\mathrm{G_1}$ and $\mathrm{G_2}$, respectively. Previous link stealing attack methods \cite{DBLP:conf/uss/HeJ0G021} used traditional models like MultiLayer Perceptron (MLP) \cite{rosenblatt1958perceptron}, which cannot handle variable-length features. Therefore, to perform attacks on these two datasets, two separate attack models need to be trained: $\operatorname{Pre_1}(\mathrm{G_1})$ and $\operatorname{Pre_2}(\mathrm{G_2})$.

LLMs were initially proposed for natural language processing tasks, and one of their advantages is the ability to handle variable-length text \cite{DBLP:journals/corr/abs-2303-18223}. This means that when facing with posterior probabilities of different dimensions, LLMs can simultaneously perform feature extraction to complete the link stealing attack task. Consequently, attackers can complete the link stealing tasks on both $\mathrm{G_1}$ and $\mathrm{G_2}$ through a single LLM, represent as $\operatorname{LLM}(\mathrm{G_1}\ or\ \mathrm{G_2})$.

In addition, let $\mathrm{n}$ denote the number of node pairs in $\mathrm{G_1}$, and $\mathrm{k}$ denote the number of node pairs in $\mathrm{G_2}$. By merging $\mathrm{G_1}$ and $\mathrm{G_2}$ for LLM training, there are $\mathrm{n+k}$ node pairs available for fine-tuning, which exceeds the node pairs $\mathrm{n}$ or $\mathrm{k}$ in a single dataset. Therefore, multiple datasets provide more training data for an LLM attack model. An LLM trained on multiple datasets is likely to have better link stealing attack performance compared to one trained on a single dataset.

\section{Experimental Evaluations}
\subsection{Experimental Setup}

\subsubsection{Datasets}
To assess the effectiveness of our proposed approach, we performed experiments on four citation datasets: Cora, Citeseer, Pubmed \cite{DBLP:conf/iclr/KipfW17}, and Ogbn-arxiv \cite{zheng2021grb}. Cora, Citeseer, and Pubmed are widely used for node categorization tasks. Ogbn-arxiv is a large dataset with $169$,$343$ nodes and $1,166,243$ links. Since our paper focuses on link stealing attacks using large language models, we include this large dataset in our evaluation. Table \ref{dataset} provides an overview of the dataset statistics.

\begin{table}[htbp]
  \centering
  \caption{Dataset statistics.}
    \begin{tabular}{c|ccccm{1.3cm}<{\centering}}
    \toprule
          & \textbf{Nodes} & \textbf{Feats} & \textbf{Links} & \textbf{Classes} & \textbf{Links in White-box}\\
    \midrule
    \textbf{Cora} & $2,708$ & $1,433$ & $10,556$ & $7$ &$2,000$\\
    \textbf{Citeseer} & $3,327$ & $3,703$ & $9,228$ & $6$ &$2,000$\\
    \textbf{Pubmed} & $19,717$ & $500$   & $88,651$ & $3$ &$5,000$\\
    \textbf{Ogbn-arxiv} & $169,343$ & $128$   & $1,166,243$ & $40$ &$30,000$\\
    \bottomrule
    \end{tabular}%
  \label{dataset}%
\end{table}%

\subsubsection{Dataset Configuration}
In white-box attacks, we suppose the attacker knows various numbers of links, depending on the dataset's size. Table \ref{dataset} shows that for the four datasets, we assume the attacker knows $2,000$, $2,000$, $5,000$, and $30,000$ links, respectively. To ensure fairness, we randomly select an equal number of unlinked edges for training.

\begin{table*}[htbp]
  \centering
  \caption{Results of different link stealing attack methods in the white-box setting.}
    \begin{tabular}{c|cccc|cccc}
    \toprule
          & \multicolumn{4}{c|}{\textbf{Accuracy}} & \multicolumn{4}{c}{\textbf{F1}} \\
\cmidrule{2-9}          & \textbf{Cora} & \textbf{Citeseer} & \textbf{Pubmed} & \textbf{Ogbn-arxiv} & \textbf{Cora} & \textbf{Citeseer} & \textbf{Pubmed} & \textbf{Ogbn-arxiv} \\
    \midrule
    \textbf{Feature \cite{DBLP:conf/uss/HeJ0G021}} & 80.51±0.29 & 77.84±0.36 & 83.65±0.18 & 81.08±0.24 & 80.78±0.36 & 78.50±0.39 & 83.66±0.23 & 81.42±0.16 \\
    \textbf{PP \cite{DBLP:conf/uss/HeJ0G021}} & 87.29±0.08 & 90.91±0.09 & 82.97±0.14 & 90.85±0.05 & 88.03±0.05 & 91.41±0.08 & 82.68±1.21 & 91.11±0.04 \\
    \textbf{PP+Feature} & 87.83±0.15 & 85.94±0.19 & 86.65±0.28 & 91.18±0.04 & 87.92±0.17 & 86.04±0.15 & 86.71±0.29 & 91.36±0.03 \\
    \textbf{LLM (Our)} & \textbf{90.34±0.08} & \textbf{93.26±0.04} & \textbf{94.08±0.02} & \textbf{97.48±0.08} & \textbf{90.19±0.08} & \textbf{93.34±0.04} & \textbf{94.08±0.02} & \textbf{97.49±0.09} \\
    \bottomrule
    \end{tabular}%
  \label{effect_white}%
\end{table*}%

\subsubsection{Models}
In this paper, we introduce Large Language Models to enhance the capability of link stealing attacks in Graph Neural Networks. In our foundational experiments, we utilize Vicuna-7B \cite{DBLP:conf/nips/ZhengC00WZL0LXZ23} as the architecture for the LLM and Graph Convolutional Network (GCN) \cite{DBLP:journals/corr/KipfW16} as the architecture for the GNN. Vicuna-7B is a common model in the LLM domain, while GCN is a widely used architecture in GNN research. To investigate the impact of different LLMs on the performance of link stealing attacks, we also incorporated Vicuna-13B \cite{DBLP:conf/nips/ZhengC00WZL0LXZ23} and LongChat \cite{longchat2023} as LLM models. Furthermore, to examine the attack performance on various GNN architectures, we included Graph Attention Networks (GAT) \cite{DBLP:conf/iclr/VelickovicCCRLB18} and GraphSAGE (SAmple and aggreGatE) \cite{DBLP:conf/nips/HamiltonYL17} models. Through experiments conducted across various models, we have demonstrated the generalizability of our proposed method.

\subsubsection{Evaluation Metric}
In this paper, we use \textbf{Accuracy} and \textbf{F1 score} to measure how well our model performs on the datasets. Accuracy shows the overall correctness of our model's predictions; it's the ratio of correct predictions to the total number of predictions. The F1 score is another metric we use, providing a blend of precision and recall, which gives us a balanced view of our model's performance, especially when dealing with imbalanced datasets. Precision indicates how many positive predictions are correct, while recall shows how many actual positives our model identifies. The F1 score combines both metrics to provide a more rounded understanding of our model's effectiveness.

\subsection{Evaluation on White-box Setting}
In the white-box setting, using some of the link data presented in Table \ref{dataset}, we fine-tune the LLM to develop models capable of performing link stealing attacks. This section explores in detail the performance of LLM in executing link stealing attacks in the white-box setting.

\subsubsection{Effectiveness of the proposed methodology}
To verify the effectiveness of our proposed method, we conducted experiments on the Cora, Citeseer, Pubmed, and Ogbn-arxiv datasets separately. The experimental results are shown in Table \ref{effect_white}. In these results, \emph{Feature} denotes that the attacker uses node features for the attack, and \emph{PP} denotes that the attacker uses posterior probabilities for the attack. In the white-box attack scenario in \cite{DBLP:conf/uss/HeJ0G021}, the attacker inputs node features or posterior probabilities into a MultiLayer Perceptron (MLP) \cite{rosenblatt1958perceptron} to train the attack model. \emph{Feature+PP} combines node features and posterior probabilities using MLP for the attack. \emph{LLM (Our)} refers to the link stealing attack method proposed in this paper, where we design prompts to fine-tune the LLM.

\begin{figure}[tbp]
  \centering
  \subfigure[Accuracy]{
  \begin{minipage}[b]{0.23\textwidth}
    \includegraphics[width=\textwidth]{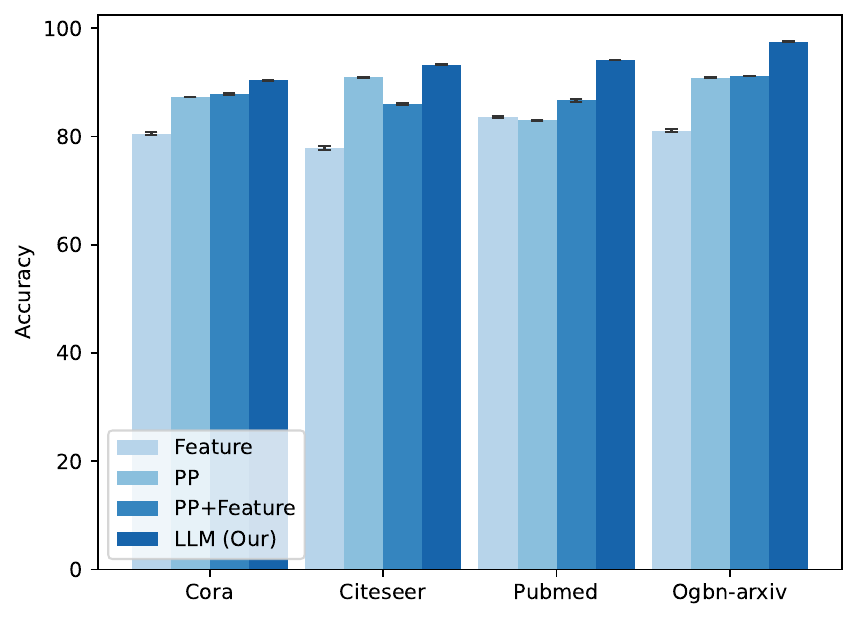}
    % \caption{Base}
  \end{minipage}
  }
  \subfigure[F1]{
  \begin{minipage}[b]{0.23\textwidth}
    \includegraphics[width=\textwidth]{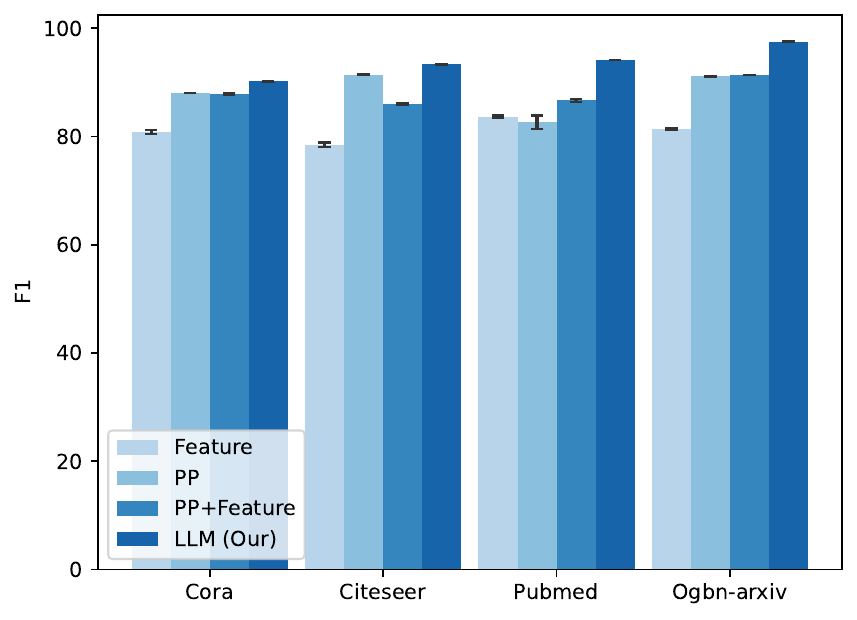}
    % \caption{Self-padding}
  \end{minipage}
  }

  \caption{Comparison of different link stealing attack methods in the white-box setting.}
  \label{com_all_white}
\end{figure}

As shown in Table \ref{effect_white}, our method achieves over $90\%$ accuracy and F1 score for link stealing attacks across all datasets. On Ogbn-arxiv, our method achieves $97.48\%$ accuracy and $97.49\%$ F1 score, demonstrating its effectiveness. The proposed method significantly enhances the performance of link stealing attacks. Particularly on the Pubmed dataset, our method improves both accuracy and F1 score by $8\%$ compared to the best previous method. Even on the Cora and Citeseer datasets, where the improvement is the least, our method still boosts performance by nearly $3\%$. To illustrate our method's superiority, we visualized the comparison of the methods, as shown in Fig. \ref{com_all_white}. This figure clearly reflects the superiority of our method compared to previous methods.

\begin{figure}[!htbp]
  \centering
  \subfigure[Accuracy]{
  \begin{minipage}[b]{0.45\textwidth}
    \includegraphics[width=\textwidth]{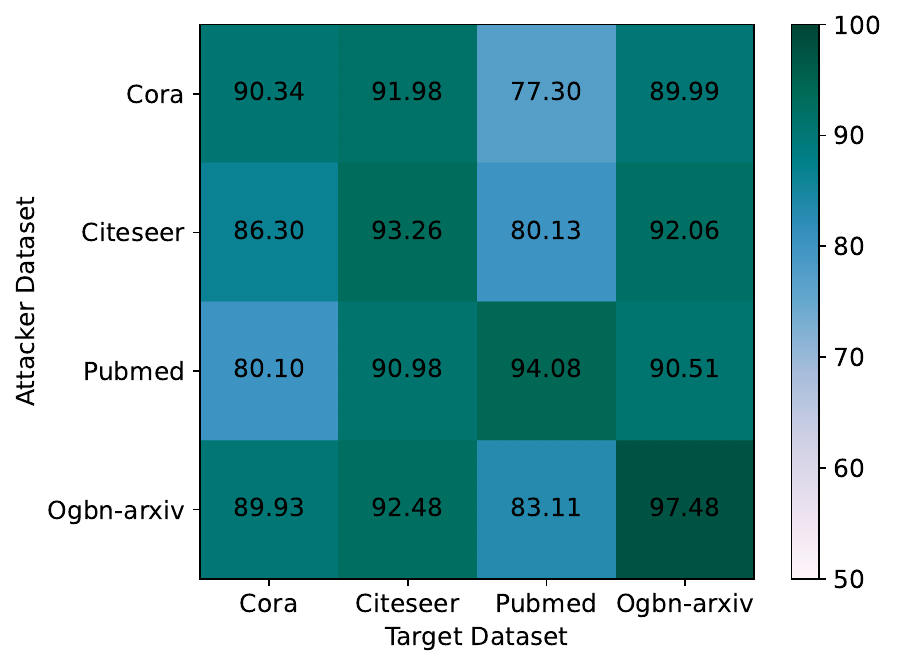}
    % \caption{Base}
  \end{minipage}
  }
  \subfigure[F1]{
  \begin{minipage}[b]{0.45\textwidth}
    \includegraphics[width=\textwidth]{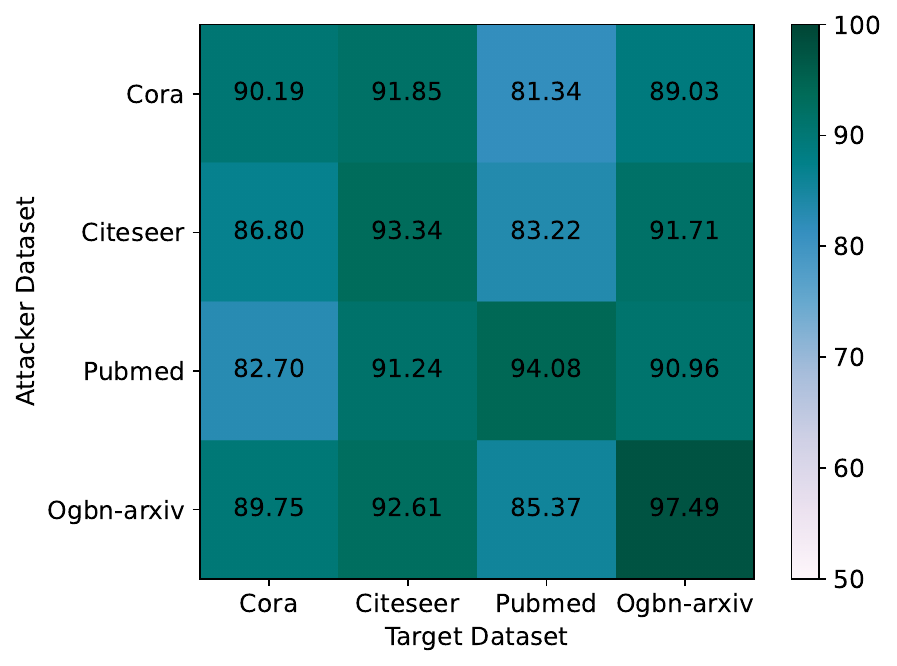}
    % \caption{Self-padding}
  \end{minipage}
  }

  \caption{Heat map of LLM performance in link stealing attacks across different datasets in the white-box setting.}
  \label{heat_map_white}
\end{figure}

\subsubsection{Training on a single dataset and attacking multiple datasets}
Here, we explore the effectiveness of LLM attack models trained on a single dataset for performing link stealing attacks on various datasets. Prior methods struggled to execute link stealing attacks across diverse datasets due to differences in posterior probability dimensions. LLM's ability to handle variable-length data addresses this challenge. We present a heatmap illustrating the performance of LLM in link stealing attacks across various datasets, as depicted in Fig. \ref{heat_map_white}.

The heatmap demonstrates that our proposed method effectively performs link stealing across different datasets. The F1 score for link stealing exceeds $80\%$ across these datasets. Furthermore, the best attack model is achieved when the attacker has access to the same links as those in the target dataset, which is represented by the diagonal line in the figure.

It is worth noting that our proposed link stealing attack method, using LLM for attacks across different datasets, often performs better than previous methods, even when the same dataset is used for both training and attacks. For instance, the LLM attack model trained on Ogbn-arxiv achieves $89.93\%$ accuracy when attacking Cora, surpassing the $87.89\%$ accuracy of the best previous attack model trained directly on Cora.

\subsubsection{Training on multiple datasets and attacking multiple datasets}
\begin{figure}[tbp]
  \centering
  \subfigure[Accuracy]{
  \begin{minipage}[b]{0.23\textwidth}
    \includegraphics[width=\textwidth]{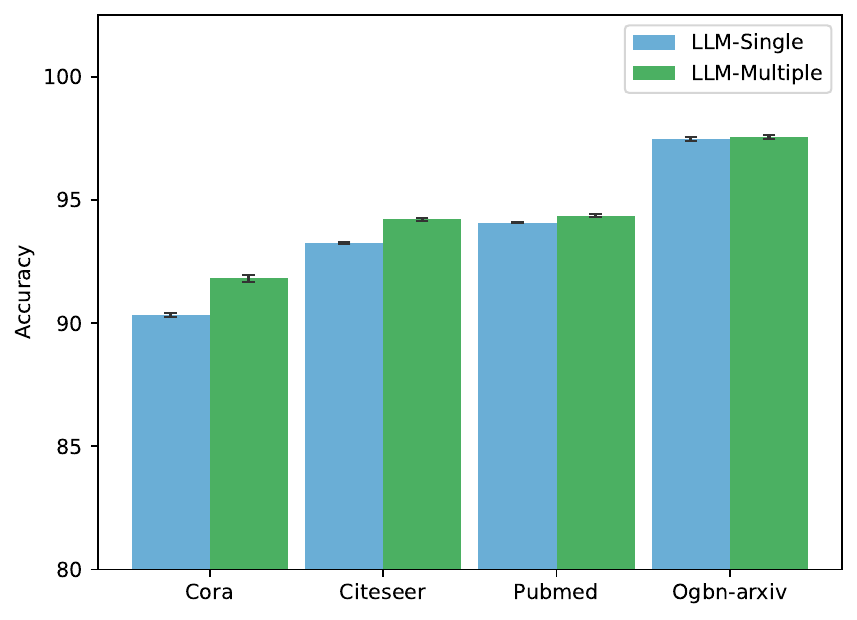}
    % \caption{Base}
  \end{minipage}
  }
  \subfigure[F1]{
  \begin{minipage}[b]{0.23\textwidth}
    \includegraphics[width=\textwidth]{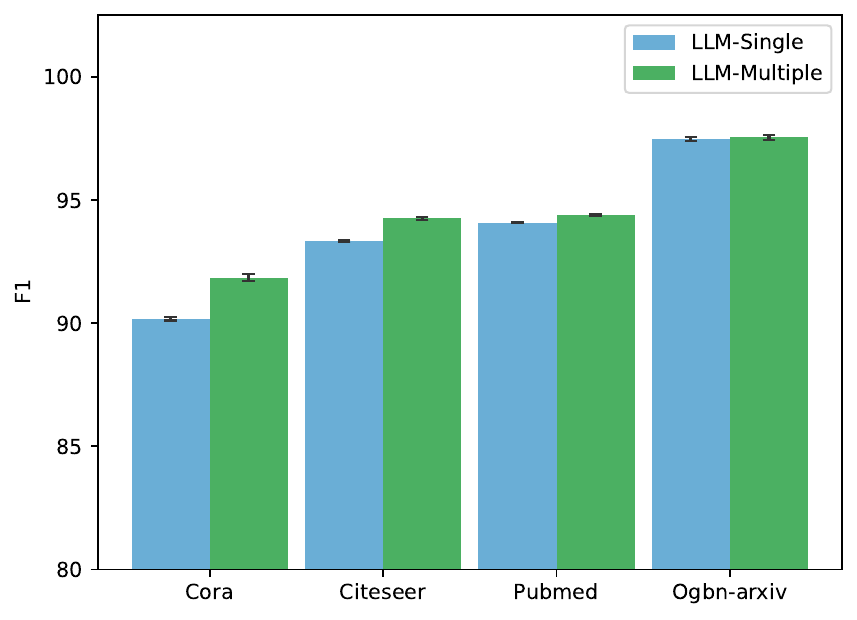}
    % \caption{Self-padding}
  \end{minipage}
  }

  \caption{Comparison of attack models trained on single vs. multiple datasets for multi-dataset attacks in the white-box setting.}
  \label{com_LLM_white}
\end{figure}

LLMs are versatile and can handle multiple tasks with one model. They can potentially manage tasks with different datasets simultaneously. We train a single LLM using data from multiple datasets for link stealing attack training. In Fig. \ref{com_LLM_white}, we compare the performance of this multi-dataset trained LLM attack model with that of the one trained on a single dataset.

The figure illustrates that the model's attack performance improves when trained on multiple datasets compared to training on a single dataset. This improvement occurs because training on multiple datasets allows the model to learn diverse dataset features simultaneously, enhancing its generalization. 

Additionally, training on multiple datasets increases the amount of training data, compensating for data scarcity, which is particularly notable in Cora and Citeseer. Table \ref{dataset} indicates that in Cora and Citeseer, the attacker has access to only 2000 linked data. The figure demonstrates a substantial enhancement in attack performance on Cora and Citeseer when using multiple datasets for training.

\begin{table*}[htbp]
  \centering
  \caption{Results of link stealing attacks by different LLMs in the white-box setting.}
    \begin{tabular}{c|cccc|cccc}
    \toprule
          & \multicolumn{4}{c|}{\textbf{Accuracy}} & \multicolumn{4}{c}{\textbf{F1}} \\
\cmidrule{2-9}          & \textbf{Cora} & \textbf{Citeseer} & \textbf{Pubmed} & \textbf{Ogbn-arxiv} & \textbf{Cora} & \textbf{Citeseer} & \textbf{Pubmed} & \textbf{Ogbn-arxiv} \\
    \midrule
    \textbf{Vicuna-7B} & 91.82±0.15 & 94.21±0.05 & 94.37±0.05 & 97.55±0.08 & 91.85±0.15 & 94.26±0.07 & 94.38±0.04 & 97.55±0.09 \\
    \textbf{Vicuna-13B} & 92.22±0.02 & 94.44±0.08 & 94.82±0.01 & 97.76±0.00 & 92.22±0.03 & 94.42±0.08 & 94.81±0.01 & 97.76±0.01 \\
    \textbf{LongChat} & 91.18±0.59 & 93.44±0.21 & 93.93±0.28 & 97.38±0.10 & 91.19±0.57 & 93.45±0.23 & 93.91±0.28 & 97.37±0.11 \\
    \bottomrule
    \end{tabular}%
  \label{LLM_white}%
\end{table*}%

\begin{table*}[htbp]
  \centering
  \caption{Results of link stealing attacks by different GNN models in the white-box setting.}
    \begin{tabular}{c|cccc|cccc}
    \toprule
          & \multicolumn{4}{c|}{\textbf{Accuracy}} & \multicolumn{4}{c}{\textbf{F1}} \\
\cmidrule{2-9}          & \textbf{Cora} & \textbf{Citeseer} & \textbf{Pubmed} & \textbf{Ogbn-arxiv} & \textbf{Cora} & \textbf{Citeseer} & \textbf{Pubmed} & \textbf{Ogbn-arxiv} \\
    \midrule
    \textbf{GCN} & 91.82±0.15 & 94.21±0.05 & 94.37±0.05 & 97.55±0.08 & 91.85±0.15 & 94.26±0.07 & 94.38±0.04 & 97.55±0.09 \\
    \textbf{SAGE} & 91.96±0.04 & 94.17±0.07 & 93.69±0.05 & 96.83±0.01 & 91.97±0.04 & 94.24±0.06 & 93.69±0.04 & 96.83±0.01 \\
    \textbf{GAT} & 93.46±0.11 & 95.87±0.03 & 94.22±0.02 & 96.91±0.01 & 93.43±0.11 & 95.92±0.03 & 94.22±0.02 & 96.90±0.00 \\
    \bottomrule
    \end{tabular}%
  \label{GNN_white}%
\end{table*}%

\subsubsection{Attacks on Other LLMs}

To explore the universality of our proposed method, we conducted link stealing attacks using different LLMs in a white-box setting. In addition to performing experiments with the widely-used Vicuna-7B LLM, we also verified our method using Vicuna-13B and LongChat. The experimental results are shown in Table \ref{LLM_white}.

It can be observed from the table that our method poses a threat to GNNs under different LLM models, achieving more than $90\%$ accuracy and F1 score across various LLMs. This indicates that the proposed method has strong universality and can successfully execute the attack under different LLM configurations.

Furthermore, the performance of link stealing attacks in the white-box setting is also influenced by the different LLM models. As can be seen from Table \ref{LLM_white}, in the case of Vicuna-13B, the link stealing attack performance is the best, achieving more than $92\%$ accuracy and F1 score. This is because, compared with the Vicuna-7B and LongChat models, Vicuna-13B has more parameters, allowing it to better fit the training data and perform more effectively in the link stealing attack task.

\subsubsection{Attacks on Other GNNs}

To explore the practicality of our proposed method, in addition to conducting experiments with the commonly used GCN as the target GNN model, we also verified our method using SAGE and GAT as the target GNN models. The experimental results are shown in Table \ref{GNN_white}.

It can be seen from the table that our method has strong practicability. Our method can effectively complete the link stealing attack under different GNN target models. Specifically, regardless of whether the target GNN model is GCN, GraphSAGE, or GAT, our method can achieve an accuracy rate and F1 score of more than $90\%$. Among them, the link stealing attack performance is the best when the target GNN model is GAT. When GAT is used as the target GNN model, even under the worst case with the Cora dataset, our method can still achieve an accuracy rate and F1 score of $93\%$. In general, although the performance of link stealing attacks varies with different target GNN models, our method can consistently complete the link stealing attack effectively for each target GNN model.

% Table generated by Excel2LaTeX from sheet '黑盒2'
\begin{table*}[htbp]
  \centering
  \caption{Results of different link stealing attack methods in the black-box setting.}
    \begin{tabular}{c|cccc|cccc}
    \toprule
          & \multicolumn{4}{c|}{\textbf{Accuracy}} & \multicolumn{4}{c}{\textbf{F1}} \\
\cmidrule{2-9}          & \textbf{Cora} & \textbf{Citeseer} & \textbf{Pubmed} & \textbf{Ogbn-arxiv} & \textbf{Cora} & \textbf{Citeseer} & \textbf{Pubmed} & \textbf{Ogbn-arxiv} \\
    \midrule
    \textbf{Feature (Mean) \cite{DBLP:conf/uss/HeJ0G021}} & 51.92±2.94 & 52.31±3.23 & 53.12±4.65 & 51.09±2.11 & 10.82±20.59 & 12.09±21.22 & 13.08±21.10 & 33.37±35.45 \\
    \textbf{Feature (Max) \cite{DBLP:conf/uss/HeJ0G021}} & 58.6  & 59.65 & 62.62 & 55.92 & 60.57  & 63.25 & 59.67 & 66.86 \\
    \textbf{PP (Mean \cite{DBLP:conf/uss/HeJ0G021})} & 81.33±11.69 & 82.19±9.57 & 74.68±9.33 & 59.29±10.90 & 76.57±27.08 & 79.97±19.70 & 72.08±24.61 & 59.64±13.54 \\
    \textbf{PP (Max) \cite{DBLP:conf/uss/HeJ0G021}} & 86.53±0.25 & 88.42±0.12 & 79.25±0.01 & 77.78±0.16 & 87.26±0.20 & 89.39±0.13 & 81.40±0.02 & 78.57±0.11 \\
    \textbf{PP+Feature (Mean)} & 59.60±12.81 & 59.54±12.45 & 61.53±15.96 & 50.78±1.30 & 26.68±35.10 & 26.79±34.18 & 30.14±39.79 & 48.52±30.11 \\
    \textbf{PP+Feature (Max)} & 75.09±0.83 & 76.34±0.06 & \textbf{84.89±0.09} & 53.13±0.10 & 70.67±0.09 & 74.24±0.08 & \textbf{84.49±0.04} & 67.19±0.04 \\
    \textbf{LINKTELLER \cite{DBLP:conf/sp/0011L0022}} & 77.75±0.24 & 81.82±0.23 & 55.46±0.11 & 72.29±0.17 & 71.38±0.38 & 79.15±2.23 & 19.71±0.36 & 61.72±0.32 \\
    \textbf{LLM (Our)} & \textbf{86.76±0.22} & \textbf{90.49±0.06} & 79.40±1.38 & \textbf{88.25±0.02} & \textbf{87.27±0.10} & \textbf{90.97±0.06} & 81.74±1.22 & \textbf{87.61±0.03} \\
    \bottomrule
    \end{tabular}%
  \label{effect_black}%
\end{table*}%

% Table generated by Excel2LaTeX from sheet '黑盒2'
\begin{table*}[htbp]
  \centering
  \caption{The accuracy of link stealing attacks under eight different similarity metrics of previous methods.}
    \begin{tabular}{c|c|cccccccc}
    \toprule
    \textbf{Dataset} & \textbf{Method} & \textbf{Cosine} & \textbf{Euclidean} & \textbf{Correlation} & \textbf{Chebyshev} & \textbf{Braycurtis} & \textbf{Canberra} & \textbf{Cityblock} & \textbf{Sqeuclidean} \\
    \midrule
    \multirow{3}[2]{*}{\textbf{Cora}} & \textbf{Feature \cite{DBLP:conf/uss/HeJ0G021}} & 52.85 & 50.1  & 52.68 & 58.6  & 50.73 & 50    & 50    & 50.4 \\
          & \textbf{GCN \cite{DBLP:conf/uss/HeJ0G021}} & 85.84±0.34 & 86.23±0.28 & 86.20±0.14 & 85.32±0.02 & 86.53±0.25 & 52.57±0.05 & 82.70±0.35 & 84.60±0.71 \\
          & \textbf{Feature+GCN} & 75.09±0.83 & 50.08±0.00 & 73.94±0.76 & 74.97±0.09 & 51.14±0.01 & 50.00±0.00 & 50.00±0.00 & 50.39±0.01 \\
    \midrule
    \multirow{3}[2]{*}{\textbf{Citeseer}} & \textbf{Feature \cite{DBLP:conf/uss/HeJ0G021}} & 53.23 & 50.28 & 53.2  & 59.65 & 51.23 & 50.1  & 50.13 & 50.65 \\
          & \textbf{GCN \cite{DBLP:conf/uss/HeJ0G021}} & 83.93±0.18 & 87.17±0.05 & 88.42±0.12 & 84.33±0.01 & 86.23±0.14 & 59.04±0.27 & 87.12±0.19 & 81.23±0.28 \\
          & \textbf{Feature+GCN} & 73.69±0.44 & 50.28±0.00 & 73.01±0.39 & 76.34±0.06 & 51.65±0.00 & 50.05±0.00 & 50.13±0.00 & 50.65±0.00 \\
    \midrule
    \multirow{3}[2]{*}{\textbf{Pubmed}} & \textbf{Feature \cite{DBLP:conf/uss/HeJ0G021}} & 56.47 & 50.02 & 55.31 & 62.62 & 50.27 & 50    & 50    & 50.26 \\
          & \textbf{GCN \cite{DBLP:conf/uss/HeJ0G021}} & 75.57±0.23 & 79.25±0.01 & 78.54±0.08 & 77.78±0.12 & 77.78±0.12 & 51.77±0.02 & 79.10±0.16 & 77.27±0.17 \\
          & \textbf{Feature+GCN} & 84.81±0.08 & 50.01±0.00 & 84.89±0.09 & 70.34±0.01 & 51.97±0.04 & 50.00±0.00 & 50.00±0.00 & 50.23±0.01 \\
    \midrule
    \multirow{3}[2]{*}{\textbf{Ogbn\_arxiv}} & \textbf{Feature \cite{DBLP:conf/uss/HeJ0G021}} & 50.12 & 50.04 & 50.12 & 55.92 & 52.33 & 50.01 & 50.01 & 50.16 \\
          & \textbf{GCN \cite{DBLP:conf/uss/HeJ0G021}} & 72.17±0.16 & 49.26±0.00 & 77.78±0.16 & 53.23±0.01 & 61.67±0.13 & 49.97±0.00 & 60.83±0.21 & 49.35±0.08 \\
          & \textbf{Feature+GCN} & 50.17±0.00 & 50.02±0.00 & 50.18±0.01 & 51.87±0.92 & 53.13±0.10 & 50.00±0.00 & 50.00±0.00 & 50.12±0.00 \\
    \bottomrule
    \end{tabular}%
  \label{eight}%
\end{table*}%

\subsection{Evaluation on Black-box Setting}
In the black-box setup, we fine-tune the LLM to determine whether nodes belong to the same class, enabling it to perform link stealing attacks. Here, we explore the LLM's link stealing attack performance in the black-box setting in detail.

\subsubsection{Effectiveness of the proposed methodology}
To verify the effectiveness of our proposed method, we conducted experiments comparing our approach with the methods presented in \cite{DBLP:conf/uss/HeJ0G021, DBLP:conf/sp/0011L0022}. Studies \cite{DBLP:conf/uss/HeJ0G021, DBLP:conf/sp/0011L0022} investigate link stealing attacks in black-box settings, and their methods are popular and effective in current research. Table \ref{effect_black} shows the experimental results. In \cite{DBLP:conf/uss/HeJ0G021}, they use eight similarity metrics to measure whether nodes have links. In the table, \emph{Mean} indicates the mean attack value of the eight similarity metrics, and \emph{Max} indicates the maximum attack value of the eight similarity metrics.

\begin{figure}[tbp]
  \centering
  \subfigure[Accuracy]{
  \begin{minipage}[b]{0.23\textwidth}
    \includegraphics[width=\textwidth]{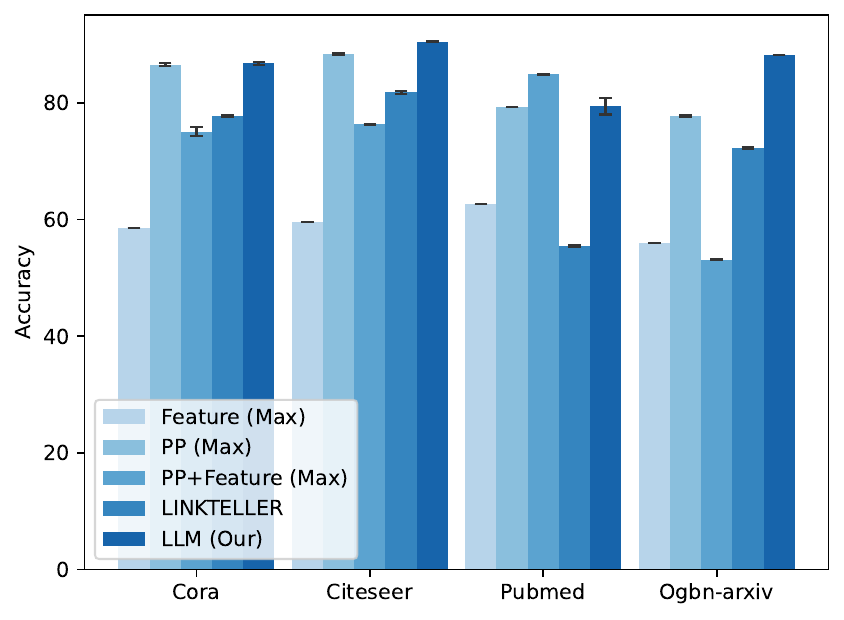}
    % \caption{Base}
  \end{minipage}
  }
  \subfigure[F1]{
  \begin{minipage}[b]{0.23\textwidth}
    \includegraphics[width=\textwidth]{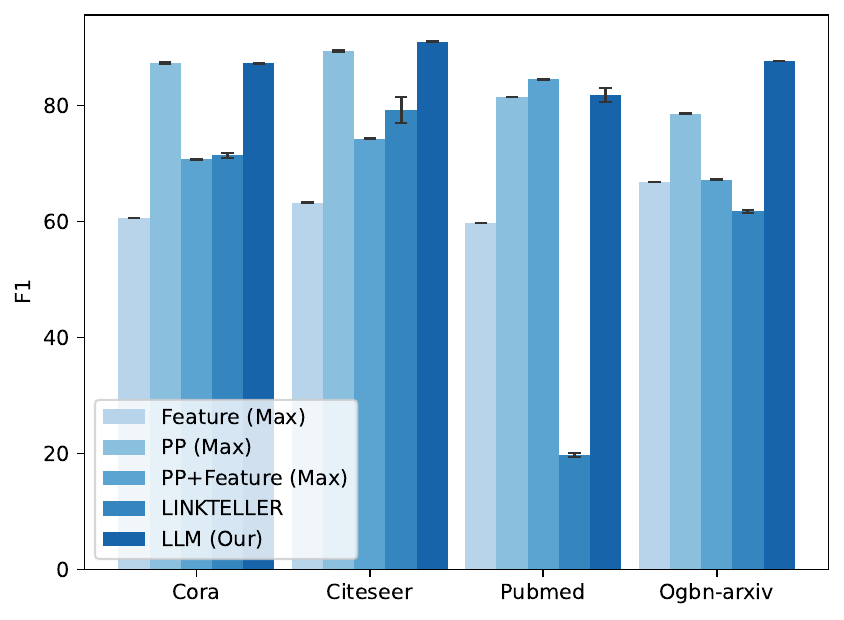}
    % \caption{Self-padding}
  \end{minipage}
  }

  \caption{Comparison of different link stealing attack methods in the black-box setting.}
  \label{com_all_black}
\end{figure}

From the table, it's evident that our method effectively performs stealing attacks in the black-box scenario. Our approach consistently achieves an accuracy and F1 score of $84\%$ across various conditions. In most cases, our method outperforms previous approaches. Particularly on Ogbn-arxiv, we observe an improvement of nearly $9\%$ compared to the best previous method. To illustrate our method's superiority, we visualized the comparison of the methods, as shown in Fig. \ref{com_all_black}.

From Table \ref{effect_black}, it is evident that in PubMed, our method slightly underperforms compared to the \emph{PP+Feature (Max)} approach. However, this method is not universally effective as it exhibits poor performance in Cora, Citeseer, and Ogbn-arxiv datasets. Moreover, \emph{Max} denotes the method that achieves the highest score among the eight similarity metrics used to assess link stealing attacks. The table reveals a significant $23\%$ point difference between the \emph{Max} and \emph{Mean} methods, indicating the instability of this approach. Due to the large variance in the Mean-related methods, we did not represent them in Fig. \ref{com_all_black}.

To further elucidate the instability of previous methods for the readers, we present the detailed accuracy of previous methods for link stealing attacks across eight different similarity metrics in Table \ref{eight}. It can be observed that, under various metrics, the performance of link stealing attacks varies significantly. For example, the  \emph{PP+Feature} method shows a $34\%$ difference in performance when measuring link stealing attacks in PubMed using the Cosine and Euclidean metrics. The table shows that no single metric can consistently perform well across different scenarios. Our method corrects this deficiency, achieving a precision greater than $84\%$ and an F1 score on all datasets, which proves to be a stable and universal approach.

% Table generated by Excel2LaTeX from sheet '黑盒2'
\begin{table*}[htbp]
  \centering
  \caption{Results of link stealing attacks by different LLMs in the black-box setting.}
    \begin{tabular}{c|cccc|cccc}
    \toprule
          & \multicolumn{4}{c|}{\textbf{Accuracy}} & \multicolumn{4}{c}{\textbf{F1}} \\
\cmidrule{2-9}          & \textbf{Cora} & \textbf{Citeseer} & \textbf{Pubmed} & \textbf{Ogbn-arxiv} & \textbf{Cora} & \textbf{Citeseer} & \textbf{Pubmed} & \textbf{Ogbn-arxiv} \\
    \midrule
    \textbf{Vicuna-7B} & 87.85±0.11 & 91.75±0.09 & 79.72±0.08 & 89.93±0.04 & 88.36±0.10 & 92.13±0.09 & 81.92±0.07 & 89.53±0.04 \\
    \textbf{Vicuna-13B} & 87.28±0.18 & 91.69±0.06 & 77.86±0.05 & 84.01±0.08 & 88.18±0.16 & 92.02±0.06 & 80.24±0.08 & 82.09±0.09 \\
    \textbf{LongChat} & 84.69±0.17 & 89.75±0.11 & 77.30±0.23 & 82.47±0.06 & 85.71±0.15 & 89.89±0.09 & 79.54±0.17 & 79.93±0.07 \\
    \bottomrule
    \end{tabular}%
  \label{LLM_black}%
\end{table*}%

% Table generated by Excel2LaTeX from sheet '黑盒2'
\begin{table*}[htbp]
  \centering
  \caption{Results of link stealing attacks by different GNN models in the black-box setting.}
    \begin{tabular}{c|cccc|cccc}
    \toprule
          & \multicolumn{4}{c|}{\textbf{Accuracy}} & \multicolumn{4}{c}{\textbf{F1}} \\
\cmidrule{2-9}          & \textbf{Cora} & \textbf{Citeseer} & \textbf{Pubmed} & \textbf{Ogbn-arxiv} & \textbf{Cora} & \textbf{Citeseer} & \textbf{Pubmed} & \textbf{Ogbn-arxiv} \\
    \midrule
    \textbf{GCN} & 87.85±0.11 & 91.75±0.09 & 79.72±0.08 & 89.93±0.04 & 88.36±0.10 & 92.13±0.09 & 81.92±0.07 & 89.53±0.04 \\
    \textbf{SAGE} & 87.74±0.06 & 91.27±0.02 & 78.06±0.04 & 89.34±0.01 & 87.79±0.07 & 91.66±0.03 & 80.00±0.04 & 88.96±0.01 \\
    \textbf{GAT} & 88.86±0.04 & 92.54±0.16 & 78.90±0.09 & 88.35±0.01 & 88.93±0.03 & 92.80±0.15 & 80.96±0.06 & 87.89±0.01 \\
    \bottomrule
    \end{tabular}%
  \label{GNN_black}%
\end{table*}%

\subsubsection{Training on a single dataset and attacking multiple datasets}
\begin{figure}[tbp]
  \centering
  \subfigure[Accuracy]{
  \begin{minipage}[b]{0.45\textwidth}
    \includegraphics[width=\textwidth]{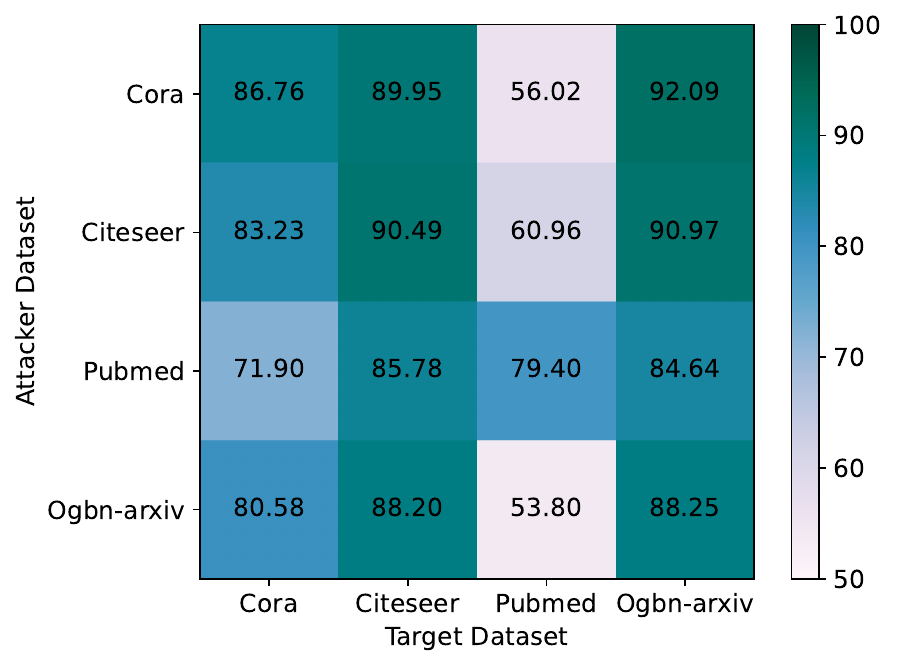}
    % \caption{Base}
  \end{minipage}
  }
  \subfigure[F1]{
  \begin{minipage}[b]{0.45\textwidth}
    \includegraphics[width=\textwidth]{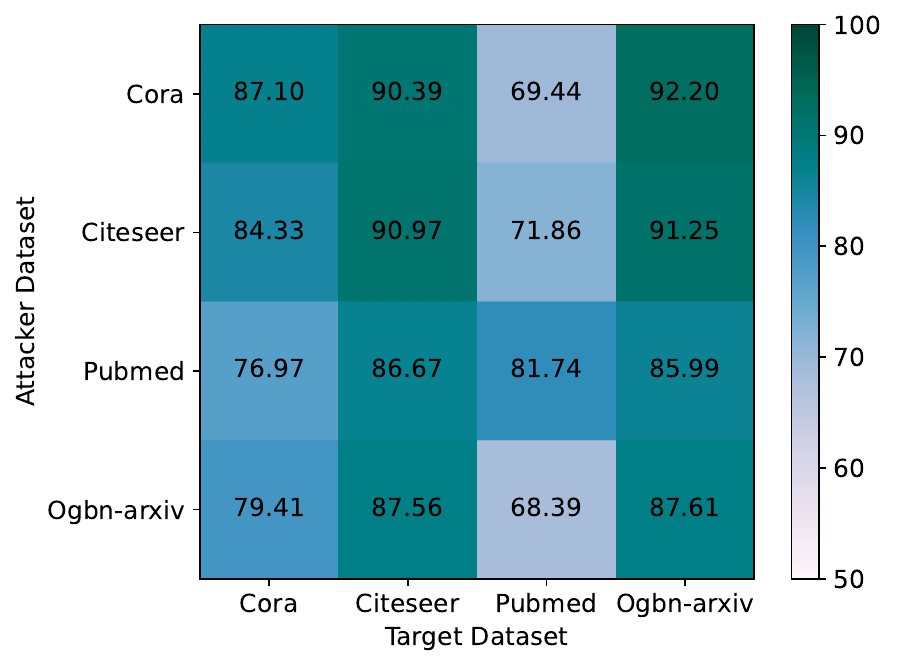}
    % \caption{Self-padding}
  \end{minipage}
  }

  \caption{Heat map of LLM performance in link stealing attacks across different datasets in the black-box setting.}
  \label{heat_map_black}
\end{figure}

In the black-box scenario, we also use LLM models trained on one dataset to perform link stealing attacks on different datasets, with the results shown in Fig. \ref{heat_map_black}. It can be seen from the figure that, in most cases, our method can still effectively perform attacks. On Citeseer and Ogbn-arxiv, our method achieves over $84\%$ attack accuracy and F1 score.

It should be noted that although our method can perform across different datasets in the black-box scenario, the effectiveness is not as good as in the white-box scenario. This is because in black-box fine-tuning, our method is only fine-tuned for tasks related to the primary objective, not specifically for the link stealing attack task. Therefore, using LLM models trained on one dataset to attack different datasets is a highly challenging cross-task, cross-dataset endeavor.

Additionally, as shown in Table \ref{dataset}, the PubMed dataset contains only three classes, so the posterior probabilities cover limited information. This limited information in the posterior probabilities makes it more difficult to simultaneously accomplish cross-task, cross-dataset tasks.

\begin{figure}[!htbp]
  \centering
  \subfigure[Accuracy]{
  \begin{minipage}[b]{0.23\textwidth}
    \includegraphics[width=\textwidth]{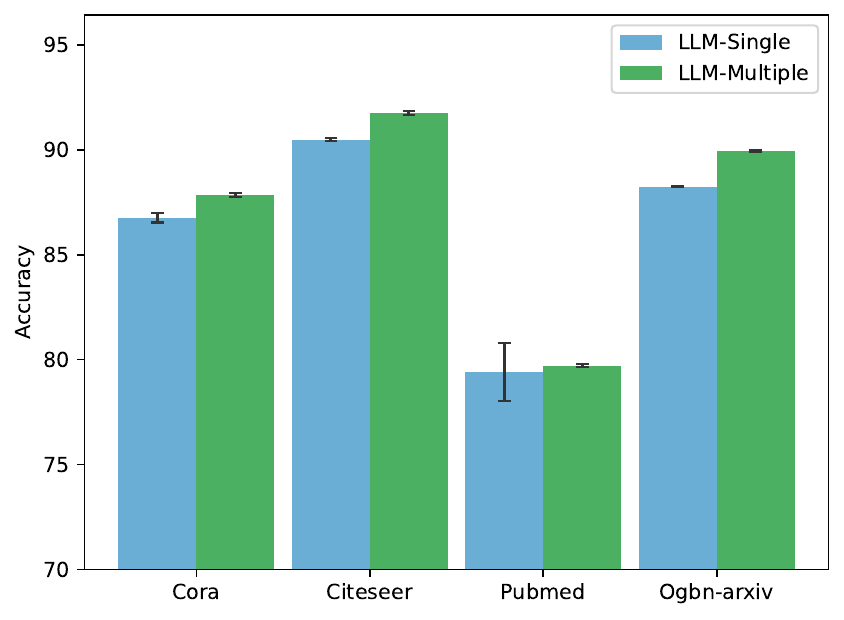}
    % \caption{Base}
  \end{minipage}
  }
  \subfigure[F1]{
  \begin{minipage}[b]{0.23\textwidth}
    \includegraphics[width=\textwidth]{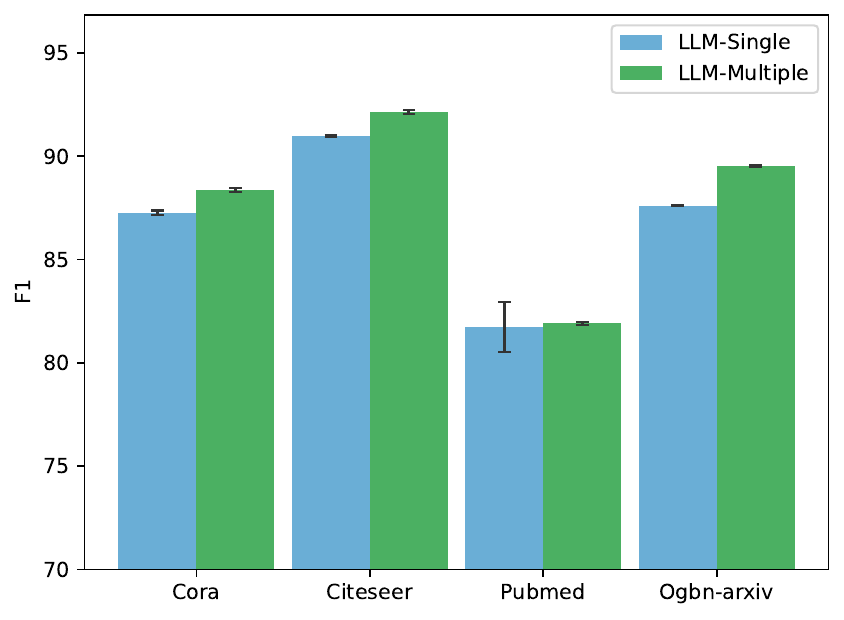}
    % \caption{Self-padding}
  \end{minipage}
  }

  \caption{Comparison of attack models trained on single vs. multiple datasets for multi-dataset attacks in the black-box setting.}
  \label{com_LLM_black}
\end{figure}

\subsubsection{Training on multiple datasets and attacking multiple datasets}
In the black-box scenario, we also use multiple datasets simultaneously to train a universal link stealing attack model, as shown in Fig. \ref{com_LLM_black}. From the figure, it can be observed that training a universal attack model with multiple datasets performs better across all datasets compared to models trained on a single dataset in the black-box scenario. This indicates that the method can leverage simultaneous training with multiple datasets in a black-box scenario to compensate for insufficient training data and enhance the model's generalization capability.

\subsubsection{Attacks on Other LLMs}

We also performed link stealing attacks using different LLMs in the black-box setting, and the experimental results are shown in Table \ref{GNN_black}. It can be seen from the table that our method is also applicable to different LLM models in the black-box setting.

It is worth mentioning that, in the black-box environment, when we fine-tune the LLM, we do not directly adopt the link stealing attack task but use similar tasks for fine-tuning. Since it is not fine-tuned directly with the link stealing attack task, the Vicuna-13B model, which can better fit the training data, does not perform as well as Vicuna-7B in this context. Of course, the LongChat model still has the worst attack performance among the three models, similar to the white-box setting. Thus, it can be seen that in the black-box setting, adopting the LLM model with moderate capability can sometimes achieve more outstanding results than the best LLM model.

\subsubsection{Attacks on Other GNNs}

We conducted link stealing attacks using various target GNN models in a black-box setting, and the results are detailed in Table \ref{GNN_black}. Despite differences in probabilities from different GNN models, which influenced attack outcomes slightly, our method consistently succeeded in executing link stealing attacks across all models. In general, our method can perform link stealing attacks on different target GNN models without having knowledge of the structure of the target GNN model, demonstrating strong practicability.

\section{Conclusion}
In this paper, we introduce large language models for enhancing link stealing attacks on graph neural networks. To the best of our knowledge, we are the first to employ LLMs for this purpose. Specifically, we develop two distinct LLM prompts to effectively combine textual features and posterior probabilities of graph nodes in both black-box and white-box scenarios. With these designed prompts, we simultaneously fine-tune the LLM using all datasets combined. Through fine-tuning, the LLM can effectively integrate the textual features of nodes, thereby enhancing the performance of link stealing attacks. Additionally, using all datasets combined to fine-tune the LLM can improve its generalization ability, enabling it to conduct link stealing attacks across different datasets with a single attack model. To validate the effectiveness of our proposed method, we conducted experiments in both white-box and black-box scenarios. Initially, we compared our method with existing popular link stealing attack methods, demonstrating the superiority of our approach. The experiment proves that our method can perform link stealing attacks on different datasets using a single attack model. We experimented with different LLMs and GNN architectures to verify the strong transferability of our method. Overall, our method surpasses existing methods and achieves state-of-the-art results.

Link stealing attacks are just one type of privacy attack on GNNs. GNNs are also susceptible to model extraction attacks, membership inference attacks, and attribute inference attacks \cite{DBLP:journals/air/GuanZZC24}. This paper demonstrates the feasibility of incorporating LLMs to enhance privacy attacks on GNNs. In the future, we will continue to explore how to utilize LLMs for other privacy attacks on GNNs.

\bibliographystyle{IEEEtran}
\bibliography{bib}

\end{document}